\documentclass{article}

\usepackage{arxiv}

\usepackage[utf8]{inputenc} 
\usepackage[T1]{fontenc}    
\usepackage{hyperref}       
\usepackage{url}            
\usepackage{booktabs}       
\usepackage{amsfonts}       
\usepackage{nicefrac}       
\usepackage{microtype}      
\usepackage{lipsum}
\usepackage{graphicx}
\usepackage{amsmath, nccmath, bm}
\usepackage{enumitem}
\usepackage{graphicx, multicol}
\usepackage{graphicx,subcaption,ragged2e}
\usepackage[utf8]{inputenc}
\usepackage{listings}
\usepackage{xcolor}
\usepackage{tgtermes}
\usepackage{amssymb}
\usepackage{float}
\usepackage{comment}
\usepackage{caption}
\usepackage{soul}

\usepackage{capt-of}
\usepackage{tabu,booktabs}
\usepackage{graphicx}
\usepackage{hyperref}
\usepackage[mathscr]{euscript}
\usepackage[outdir=./]{epstopdf}
\usepackage{natbib}
\DeclareUnicodeCharacter{FB01}{\,}
\DeclareUnicodeCharacter{0395}{$E$}
\DeclareSymbolFont{rsfs}{U}{rsfs}{m}{n}
\DeclareSymbolFontAlphabet{\mathscrsfs}{rsfs}
\graphicspath{ {./} } 

\title{A Multi-Fidelity Graph U-Net Model for Accelerated Physics Simulations}

\author{
 Rini Jasmine Gladstone \\
  Department of Civil and Environmental Engineering\\
  University of Illinois at Urbana-Champaign\\
  \texttt{rjg7@illinois.edu} \\
   \And
 Hadi Meidani \\
  Department of Civil and Environmental Engineering\\
  University of Illinois at Urbana-Champaign\\
  \texttt{meidani@illinois.edu} \\
}

\begin{document}
\maketitle
\begin{abstract}
Physics-based deep learning frameworks have shown to be effective in accurately modeling the dynamics of complex physical systems with generalization capability across problem inputs. While unsupervised networks like PINNs rely only on the underlying governing equations and boundary conditions, they are limited in their generalization capabilities. However, data-driven networks like GNN, DeepONet, Neural Operators have proved to be very effective in generalizing the model across unseen domains, resolutions and boundary conditions. But one of the most critical issues in these data-based models is the computational cost of generating training datasets. Complex phenomena can only be captured accurately using large enough deep networks that require large training datasets. Furthermore, the numerical error of the samples in the training data is propagated in the model errors, which necessitates the need for accurate data, i.e. FEM solutions on high-resolution meshes. Multi-fidelity methods offer a potential solution to reduce the training data requirements. To this end, we propose a novel GNN architecture, Multi-Fidelity U-Net, that utilizes the advantages of the multi-fidelity methods for enhancing the performance of the GNN model. The proposed architecture utilizes the capability of GNNs to manage complex geometries across different fidelity levels, while enabling flow of information between these levels for improved prediction accuracy for high-fidelity graphs while optimizing computational demands. We show that the proposed approach performs significantly better in accuracy and data requirement and only requires training of a single network compared to other benchmark multi-fidelity approaches like transfer learning models. We also present Multi-Fidelity U-Net Lite, which is a faster version of the proposed architecture, with 35\% faster training time per iteration, with 2 to 5\% reduction in accuracy. We carry out extensive validations across various simulation data to show that the proposed models surpass traditional single-fidelity GNN models in their performance, thus providing feasible alternative for addressing computational and accuracy requirements where traditional high-fidelity simulations can be time-consuming.
\end{abstract}


\section{Introduction}

Scientific modeling, governed by partial differential equations (PDEs), often relies on high-fidelity solvers like finite element and finite volume techniques. However, these methods can be computationally expensive and are not suitable for real-time physical simulations. To address this challenge, advanced deep learning techniques have been used to create neural network surrogates that can efficiently and accurately solve physical systems, providing faster results than traditional solvers. These surrogates can be trained using data-driven or physics-based methods, or a combination of both \cite{thuerey2021pbdl}. Some of the popular neural network models for this class of problems are Physics-informed neural networks (PINNs) \cite{raissi2017physics}, DeepONets \cite{lu2021learning} and Neural operators \cite{li2020neural} and Graph Neural Networks (GNNs) \cite{gladstone2023gnnbased, DBLP:journals/corr/abs-2010-03409}. There has been an explosion in recent work where these networks are used as differentiable models that can replace traditional solvers to become learned simulators in different domains and physical systems \cite{multiscale, gladstone2022fo, li2020fourier,li2020multipole}. 

Physics based neural networks are a class of neural networks trained by incorporating the physics (PDEs, boundary and initial conditions) in the loss terms  \cite{khoo2021solving, lagaris1998artificial, RAISSI2019686, raissi2017physics, sirignano2018dgm, yu2018deep}. These models can be used for solving physical systems with specific set of PDE parameters and boundary conditions (BCs) and hence has less generalization capability. There has been developments to improve the performance of PINNs for parameterized systems, where the BCs, PDE parameters and geometries can be parameterized. One of the methods is to use first order PINNs (FO-PINNs) \cite{gladstone2022fo}, where the output of the models consists of both the predicted response of the system as well as their first order spatial derivatives. 

Convolutionals Neural Networks (CNNs) \cite{tompson2016accelerating},  DeepONets and neural operators are some of the popular neural network surrogates that are fully data driven. CNNs learn the response of the system over a discretized domain by using observed data from simulations using traditional solvers \cite{wei2019physics}. While CNNs approximate functions to predict the response for a specific PDE problem with fixed boundary conditions, DeepONet approximates both linear and nonlinear operators, by taking functions as inputs, and mapping them to other functions in the output space. Neural operators are an extension to DeepONet, which also aims to directly learn the solution operator of PDEs. Both DeepONet and neural operators are independent of resolution and training grid and hence allows for zero-shot generalization to higher resolution evaluations \cite{li2020neural, lu2021learning}. 

One of the more recent developments in PDE solvers is GNNs, where the computational domain is represented as a graph thus, making the network invariant of the resolution of the data or the domain. Mesh based GNNs called MeshGraphNet \cite{multiscale} and their extensions have been successfully used for solving time dependent dynamic problems and  generalize well to different resolutions and boundary conditions.  \cite{gladstone2023gnnbased} introduced Edge Augmented GNN which is successfully used for solving time independent linear elastic problems. They introduced augmented edges to the existing graphs/meshes for faster propagation of messages across the computational domain, thus resulting in more accurate evaluation of responses with less deeper GNN networks. The U-Net architecture is also employed in GNN networks, similar to multigrid methods, and have been successfully used for solving various mesh based simulations \cite{deshpande2024magnet, gladstone2024mesh}. 

The generalization capability of a surrogate model, data-driven or physics-informed, is indicated by the stability of the model for a variety of PDE formulations and computational domains \cite{wang2021understanding, zhu2019physics}. The trained surrogate model must be able approximate the response of the physical systems for a range of boundary conditions and material domains without the need of re-training.  In this context, multi-fidelity modeling \cite{giselle2019issues} using neural networks have been proposed to improve the generalization capability of the models. However, beyond issues related to generalization, one of the most critical issues in the data driven models is the computational cost of the generation of training datasets. Complex phenomena in the physical systems are captured accurately using large enough deep networks that require large training datasets. Furthermore, the total error of the trained model is the sum of the training error of the network and the numerical error of the samples in the training set \cite{liu2022multi}. This necessitates that the training data should consist of accurate solutions, which in most cases would be FEM simulations computed on very fine meshes. However, generating such high-fidelity data is computationally very expensive. Moreover, it would require deeper neural networks and higher GPU specifications for training the models with this data. An alternative to this challenge is the use of multi-fidelity approaches that rely on low-fidelity data along with high-fidelity data for better training. There are many research works that focus on using multi-fidelity methods in neural networks to reduce the computational requirements for the training data and to enhance generalization capability of the models. A review of the works on multi-fidelity methods in neural networks is given in Section~\ref{mf_works}.

To address the challenges in the applications of GNN in physics simulations as detailed above, we propose a novel GNN architecture, Multi-Fidelity U-Net, that utilizes the advantages of the multi-fidelity methods for performance enhancement. The proposed architecture enables bi-directional information flow between graphs of different fidelities during the training and thus, improves the performance of the multi-fidelity GNN models. For this, it employs a U-Net architecture, where each level in the downward and upward parts of the U-Net architecture processes graphs of different fidelities. Here, the information flows in two directions, from high- to low-fidelity graphs and vice versa, via a coupling process, by adding the node attributes from  the lower fidelity graph to that of next higher fidelity graph through an up-sampling process and higher to lower fidelity level through down-sampling. Moreover, the output from all the levels of fidelities are used in loss function for training the models. We also present, Multi-Fidelity U-Net Lite, a faster version of the proposed architecture, where the information flow is uni-directional, from just low- to high-fidelity level and not vice versa. For this study, we have considered domain resolution as the fidelity of the model. The efficacy of the Multi-fidelity U-Net architecture is showcased through a variety of numerical examples, including displacement evaluation for 2D cantilever beams with varying geometry and boundary conditions, plane stress concentration analysis for 2D plates with different geometric variations, and finally, a large-scale 3D computational fluid dynamics (CFD) simulation using an industry-standard aerodynamics data set, containing 3D vehicle models of meshes with hundreds of thousands nodes. Moreover, we show that we get around 35\% faster training time with Multi-fidelity U-Net Lite with only 2 to 5\% decrease in accuracy for these experiments. Through these examples, we demonstrate the superior performance of the proposed models over single-fidelity GNN architectures and a benchmark multi-fidelity GNN model, transfer learning approach \cite{liu2022multi}. 

The rest of the paper is organized as follows: Section~\ref{background} gives a background on multi-fidelity methods and provides the concepts required to understand the working of GNNs along with the relevant literature review. The methodology proposed by this study is detailed in Section~\ref{methodology}. The analysis of numerical results is presented in Section~\ref{results}. This is followed by
concluding remarks and discussion of future works in Section~\ref{discussion}.

\section{Background}
\label{background}

\subsection{Multi-fidelity methods}
\label{mf_works}

The accuracy of predictions for scientific computing problems is heavily dependent on the available fidelity of the data \cite{fernandez2016review, peherstorfer2018survey}. Typically, high-fidelity models lead to more accurate evaluation of responses, but has prohibitively higher data generation and computational cost, thus limiting its applicability. Alternatively, low-fidelity models, which are more computationally affordable, are less accurate due to the simplification in terms of dimensionality reduction, linearization, use of simpler physics models or coarser domains \cite{fernandez2016review}. Multi-fidelity methods aim to bridge the gap between the accuracy and computational cost of low- and high-fidelity models by leveraging capabilities of both the fidelities in a single model. In the context of our current study, the different fidelities refer to the domain resolution, i.e., high and low-fidelity data refer to meshes with high- and low-resolution. 

There are many research works that focus on using multi-fidelity methods in neural networks to reduce the computational requirements for the training data and to enhance information gain of the models. \cite{li2023multi} uses fusing methods to integrate low-fidelity data, obtained from reduced models, with high-fidelity data, obtained from expensive simulations, to build high-performance surrogate models. \cite{meng2020composite} proposed a multi-fidelity deep neural network for function approximations and solving inverse PDEs. \cite{mahmoudabadbozchelou2021data} used multi-fidelity neural network framework to model the rheological properties of fluids, by implicitly incorporating the physics information in the form of low-fidelity data, thus enhancing the prediction capability of the model. 
\cite{pawar2022towards} used a combination of principal component analysis and multi-fidelity neural networks for predicting the flow fields of wind turbine wakes, with good prediction accuracy for 2D cross-section velocity field of the wake. \cite{zhang2021multi} and \cite{he2020multi} used multi-fidelity neural networks for combining different fidelity data and predicting the aerodynamic performance of airfoils. \cite{gladstone2021robust} used multi-fidelity variational autoencoders for generating high-resolution geometries, by using multi-fidelity data for training the model and using them for robust topology optimization problems.

In the realm of scientific machine learning, multi-fidelity physics informed neural networks have been trained using low fidelity data and evaluated on high fidelity physics \cite{chakraborty2021transfer, gorodetsky2021mfnets, meng2020composite}. Low fidelity training data improves the computational cost of data generation and model training. Some physics-based surrogate models have also generalized across multiple physical domains by applying multi-fidelity approaches to learning within a subdomain \cite{wang2021train}. Using multi-fidelity data for training can also increase the information gain of the model, thus further improving the generalization capability of the network \cite{black2022learning, chen2021learning, raissi2017inferring, zhang2020machine}. 

While the above works focus on using multi-fidelity concept on fully connected neural networks, there are a few studies that have used multi-fidelity concept in GNNs for solving PDE problems. A prominent example for this is MFT \cite{li2023multi, liu2022multi}, a multi-fidelity transfer learning approach that uses two meshes of different granularity. On the coarse mesh, approximate FEM solutions for the PDE can be computed at low cost. However, this approximates the exact solution of the PDE poorly due to the coarseness of the mesh. On the other hand, we can compute more accurate approximations of the exact PDE solutions on the fine mesh, at a very high computational cost. This however makes it unrealistic to generate enough samples to train an accurate deep learning model. MFT trains such a deep model using only a small dataset of highly accurate solutions, but leveraging the result of the training on the large training data computed on the coarse mesh, using transfer learning. The model trained on the large dataset generated on the coarse mesh is used as a starting point for the learning process on the small dataset of accurate solution. However,  the information gain from the coarse data on the network trained using the fine data is only through transfer learning and is offline in nature as they are not trained together. However, this would lead to longer training times due to multiple training of GNNs with similar architectures. In another related work, \cite{black2022learning} proposed MFGNNs, where low-fidelity projections trained using coarse data are used to inform high-fidelity modeling across arbitrary subdomains in a fine mesh represented by subgraphs. However, this method is primarily used for 2D uniform grids, limiting their extension to more complex, higher dimensional problems. Moreover, due to the need to create subdomains, it can lead to computational challenges when used for large scale problems. The latest work in this area are two GNN architectures proposed in \cite{taghizadeh2024multifidelity}, a hierarchical multi-fidelity GNN which uses a large number of low-fidelity data to train a GNN surrogate for evaluating the PDE solutions for coarse mesh. The predicted values from the coarse mesh is upsampled and fed as additional node features to the high-fidelity graphs used for training a separate GNN surrogate for the fine meshes. However, this method also requires separate training of low- and high-fidelity GNNs which increases the total training time.

In this paper, we introduce a novel method to combine multi-fidelity modeling in GNN architecture as well as the training process to enhance the performance of GNNs for solving PDE problems, while addressing the challenging in the existing works. This method extends the application of GNNs to PDE problems
with complex and irregular meshes, while optimizing the computational cost in data generation and model training.

\subsection{Graph Neural Networks}
\label{GNN}

Graph Neural Networks (GNNs) are a class of deep learning methods that operate on graph structures consisting of nodes and edges \cite{zhou2020graph}. GNNs  use the concept of message passing between neighboring nodes in the graph to model the interdependence of features at various nodes. The GNN architectures and their performance vary primarily due to their algorithms for message passing and propagation between the nodes. Some of the popular GNN architectures are graph convolutional networks \cite{kipf2016semi}, GraphSAGE \cite{hamilton2017inductive}, Graph attention networks \cite{velivckovic2017graph}, Graph transformer networks \cite{yun2019graph}, Graph Networks \cite{battaglia2018relational} etc.  Due to their ability to handle complex and irregular domains and their superior performance in the predictive power, GNNs have found applications in a variety of problems, where the data can be represented in graph structures, such as in particle physics \cite{shlomi2020graph}, high energy physics detectors \cite{ju2020graph}, power systems \cite{donon2019graph}, etc. Additionally, GNN frameworks have been able  to simulate complex physical domains involving fluids, and rigid solids, and deformable materials interacting with one another \cite{pmlr-v119-sanchez-gonzalez20a}. In MeshGraphNets, the pioneering work for using GNN for solving PDE problems, the mesh physical domain is represented using a mesh, which is basically a graph on which GNNs learn to predict physics (see e.g., \cite{DBLP:journals/corr/abs-2010-03409}).

For this study, we consider a GNN network with the the same high-level network architecture as the MeshGraphNet  \cite{DBLP:journals/corr/abs-2010-03409} whose building blocks are: (i) an encoder, (ii) $m$ GN blocks, and (iii) a decoder. The mesh is  converted into a graph structure by identifying the vertices of the mesh as nodes and the connections between the vertices in the mesh as the edges.  The node attributes contain the necessary information for GNN to learn the response of the physical system. This could include nodal coordinates, boundary or interior nodes, details of boundary conditions etc. The edge attributes usually contain information about the relative position and distance between the nodes. 

Once the graph is generated, the node attributes and edge attributes are encoded into a latent space through an encoder. The encoded nodes and edges attributes are then passed through $m$ GN blocks and updated. Each GN block consists of an edge update module and a node update module. Let $\bm{n}'_i$ be the encoded node attribute vector of node $i$ and $\bm{n}'_j$ be the encoded node attribute vector for node $j$, such that $j \in N(i)$, where $N(i)$ denotes the neighborhood of node $i$. Let $\bm{e}'_{ij}$ be the encoded edge attribute vector for the edge between $i$ and $j$. The edge update module, $\chi$, is a MLP that receives the attributes of the nodes connecting the edge along with the edge attributes and returns the updated attributes, i.e. $\bm{e}'_{ij} = \chi(\bm{e}'_{ij})$. 

The node update module consists of three MLPs, namely $\phi$, $\gamma$ and $\beta$. Specifically, the encoded node attributes of node $i$, denoted by $\bm{n}'_i$, are updated as follows \cite{gladstone2023gnnbased}:
\begin{equation}
    \begin{aligned}
    \label{eq:node_update}
    & \bm{m}_{ij} = \phi (\bm{n}'_i, \bm{n}'_j, \bm{e}'_{ij}),\\
    & \bm{n}'_{i} \leftarrow \gamma \left(\bm{n}'_i, \frac{1}{M}\sum_{j \in N(i)}\bm{m}_{ij}\right),\\
    & \bm{n}'_{i} \leftarrow \bm{n}'_{i} + \beta (\bm{n}'_{i}),
    \end{aligned}
\end{equation}
where $\bm{m}_{ij}$ is the message passed between the nodes $i$ and $j$, calculated as a function of the attributes of the connected nodes, $\bm{n}'_i$ and $\bm{n}'_j$, as well as the attributes of the connecting edge, denoted by $\bm{e}'_{ij}$. Finally, the updated node attributes, $\bm{n}'$, is passed through a decoder to return the response of the physical systems. Both the encoder and decoder functions are  MLPs.

\section{Methodology}
\label{methodology}

In this study, we propose a multi-fidelity framework based on the Graph U-Net, called Multi-Fidelity U-Net, for the training of mesh-based GNN networks using multi-fidelity data from simulations. The proposed architecture is characterized by bi-directional flows of information between graphs of different resolution. That is,  the information is transferred from high- to low-fidelity graphs and vice versa. We also propose a lighter version of the proposed method, called the Multi-Fidelity U-Net Lite. That is, the information flow is uni-directional from lower to higher fidelity graphs only. This results in faster training. An illustration of these methods can be found in Fig. \ref{fig:methodology_mf_a} and \ref{fig:methodology_mf_b} respectively. The subsequent sections provide a detailed explanation of the proposed architectures.

\subsection{Multi-Fidelity U-Net (MF-UNet)}
\label{mfgunn_method}

\begin{figure}
     \begin{subfigure}[b]{0.98\textwidth}
         \centering
         \includegraphics[width=\textwidth]{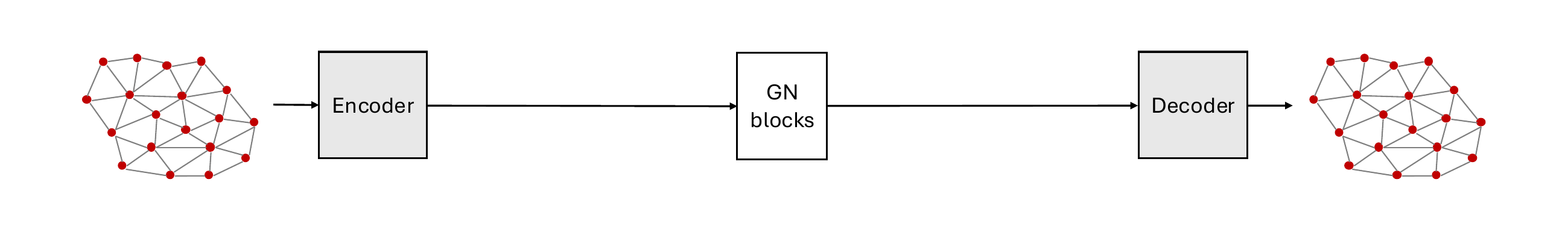}
         \caption{Single Fidelity GNN}
         \label{fig:sf_method}
     \end{subfigure}
    \hfill
     \begin{subfigure}[b]{0.98\textwidth}
         \centering
         \includegraphics[width=\textwidth]{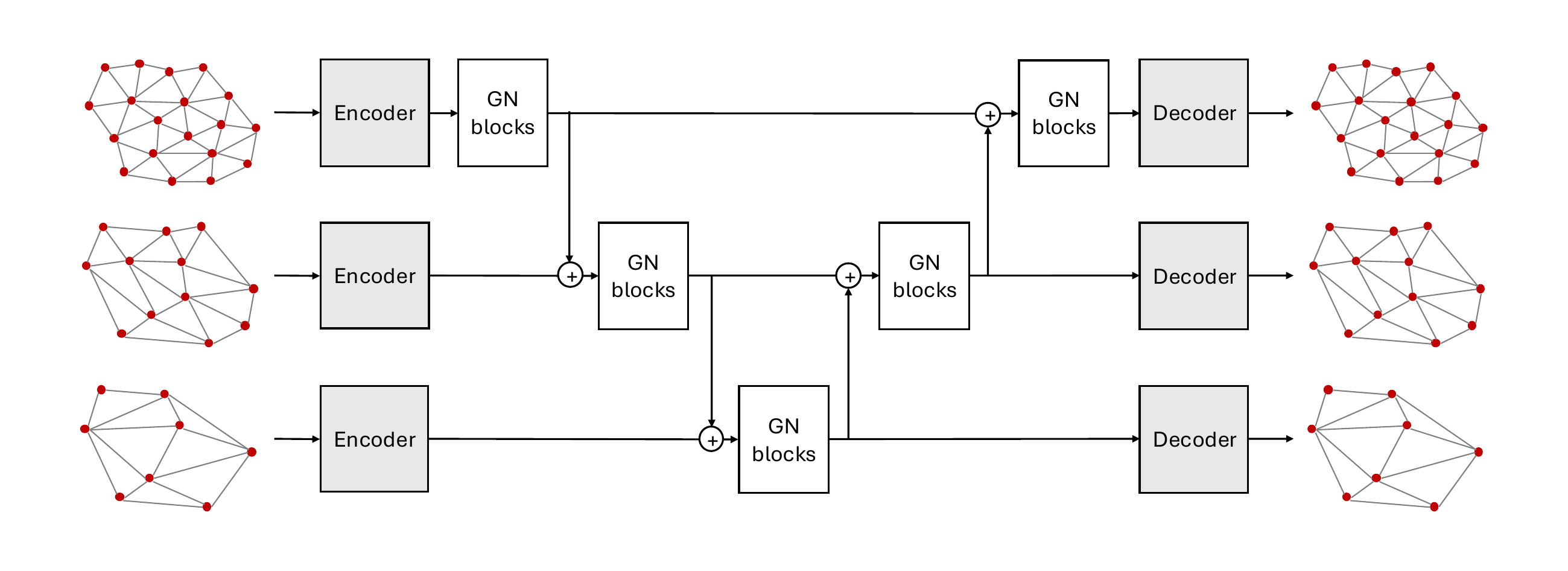}
         \caption{Multi-Fidelity U-Net}
         \label{fig:mf_method}
     \end{subfigure}
     \caption{An overview of  (a) a single-fidelity GNN architecture; and  (b) our proposed multi-fidelity U-Net architecture, where information flows in two directions:  both  from higher- to lower-fidelity levels and also from lower- to higher-fidelity. The downward flow from high- to low-resolution level is carried out by adding updated node attributes from the higher-fidelity level to the encoded node attributes of the lower-fidelity level. The upward flow involves adding updated node attributes (from the last GN block) of the lower-fidelity to the updated node attributes of the next higher-fidelity, before it is passed through another set of GN blocks. As the number of nodes are different for graphs of different fidelity, down-sampling and up-sampling operations are carried out based on the nodal distance to calculate nodal information between different levels. Graphs from all the levels of fidelity share the same encoder, decoder and GN blocks.}
    \label{fig:methodology_mf_a}
\end{figure}

In this section, we present the proposed Multi-Fidelity U-Net architecture using multi-fidelity data for GNN models. Here, the predictive power of the model for the high-fidelity simulations is enhanced by enabling a bi-directional flow of nodal information between graphs of different  resolutions. An overview of the proposed architecture can be found in Fig.\ref{fig:methodology_mf_a}. Without the loss of generality, let us consider three levels of fidelity, namely High Resolution (HR), Medium Resolution (MR) and Low Resolution (LR) graphs. It should be noted that in general two or more fidelity levels can be considered for this architecture, and  as the number of fidelity levels  increases, the computational complexity of the model increases, as well. The GNN architecture to process the graphs at each level is similar to that of single-fidelity GNN, consisting of an encoder, a series of GN blocks and finally a decoder. 

Here, the processing of information starts at the highest fidelity level, with the HR graphs. The node attributes of the HR graph, $\bm{n}_1$ are encoded using an encoder, and passed through a few GN blocks to get intermediate update of node attributes,  $\bm{n}'_1$. These are, then, added to the encoded node attributes from the next lower fidelity level data, MR graphs, and then passed through a few GN blocks to get the intermediate update of the node attributes at medium-fidelity level, $\bm{n}'_2$. Similarly, $\bm{n}'_2$ is passed to the next lower level of fidelity, by adding it to the encoded node attributes of LR graphs, $\bm{n}_3$. These are then passed through all the GN blocks to get the updated node attributes for the LR graphs, which is fed into the decoder to get the predicted response of the LR graphs,  $\hat{\bm{u}}_3$. These steps constitute the flow of information from high- to low-fidelity level. The information flow in the opposite direction (low- to high-fidelity level) starts with adding the  updated node attributes from the LR graph to the intermediate node attributes of MR graph, $\bm{n}'_2$, and they are passed through a series of GN blocks to get the final updated node attributes for the MR graphs. This is then fed to the decoder network to get the predicted response for MR graph, $\hat{\bm{u}}_2$. The updated node attributes from this level are also passed to the high-fidelity level, by adding them to the intermediate node attributes of HR graph, $\bm{n}'_1$, which are passed through the GN blocks and fed to the decoder to obtain the predicted response for the HR graph, $\hat{\bm{u}}_1$. Thus, information is passed from high- to medium- to low-fidelity levels and vice versa to enrich the model, via coupling of the node attributes. Here, coupling is defined as the process of adding the node attributes from  the lower fidelity graph to that of next higher fidelity graph as well as higher to lower fidelity level.

It is to be noted that, as the number of nodes are different for LR, MR and HR  graphs, an up-sampling and down-sampling process is done to enable the coupling process. For this, we use $k$-nearest nodes based on Euclidean distance between the nodes to map every node in the LR graph to $k$ nodes in the MR graphs and similarly, that in the MR graph to $k$ nodes in the HR graphs. Thus for the up-sampling process, the node attributes from the LR graph is added to the node attributes of the k-nearest nodes in the MR graph and the node attributes from the MR graph is added to the node attributes of the k-nearest nodes in the HR graph. The impact of the added node attributes from the lower level of fidelity is controlled by a weight parameter, $\beta_1$, which is learned during the training of the model. Similarly, for the down-sampling process, when the node attributes are passed from higher to lower level of fidelity, a mean of the node attributes of the $k$ nearest nodes from the higher resolution graph is added to the node attributes of the lower resolution graph, and its impact is controlled by a learned weight parameter, $\beta_2$. Additionally, the encoder, GN blocks and the decoder of all the levels of fidelity have shared parameters during the training of the model. 

Unlike the transfer learning method, which uses large number of low-fidelity data for training using LR graphs and small number of high-fidelity data for training MR or HR graphs, MF-UNet uses the same number of low-, medium- and high-fidelity data for training, as these graphs are trained together. Thus, during data generation, for every simulation, we simultaneously generate low-, medium and high-resolution meshes and corresponding solutions. A data pre-processing step is carried out by extracting relevant features such as nodal coordinates and boundary conditions from these simulation results from meshes of all the resolutions. The mesh data are converted into graphs, where nodes represent mesh points and edges represent the connectivity between these points. We also create edge attributes which contain relative distances between the nodes in each direction as well as the Euclidean distance between them. The model is trained by using the predicted responses from all the three levels of fidelity,  $\hat{\bm{u}}_1$, $\hat{\bm{u}}_2$ and $\hat{\bm{u}}_3$. The respective ground truth for these graphs, $\bm{u}_1$, $\bm{u}_2$ and $\bm{u}_3$ from the finite element simulations are used for calculating the loss function for training the model.


\subsection{Multi-Fidelity U-Net Lite (MF-UNet Lite)}
\label{coupled_mf}

\begin{figure}
     \centering
     \includegraphics[width=\textwidth]{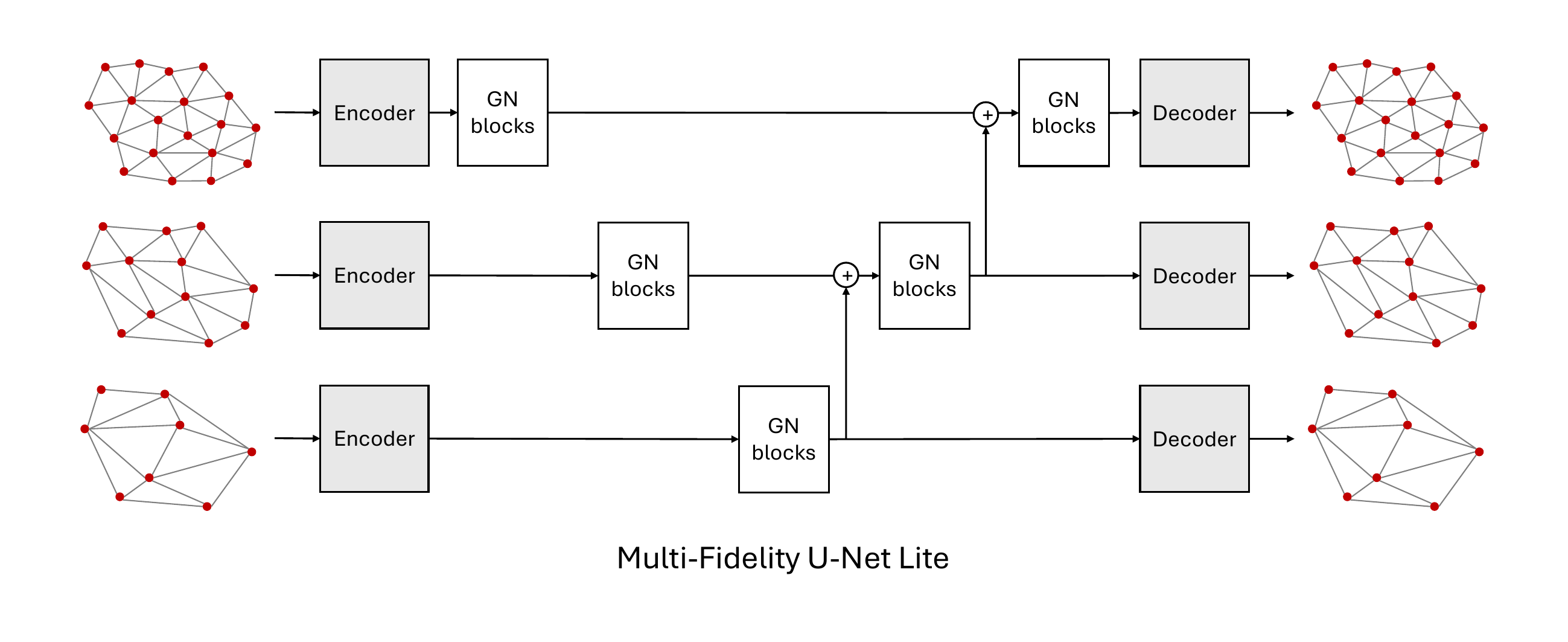}
     \caption{An overview of the Multi-Fidelity U-Net Lite architecture, which is similar to the Multi-Fidelity U-Net architecture, with the information flowing in one direction - from lower- to higher-fidelity levels. Similarly to Multi-Fidelity U-Net, the upward flow of information is through the coupling between the updated node attributes from the last GN block on the lower resolution graph and the updated node attributes of the next higher resolution graph, before its passed through another set of GN blocks. We  use up-sampling operation based on nodal distance, to pass nodal information between the fidelity levels. Moreover, encoder, decoder and GN blocks share parameters across different levels.}
     \label{fig:methodology_mf_b}
\end{figure}

In this section, we present a lighter version of the proposed multi-fidelity architecture, Multi-Fidelity U-Net Lite, where the flow of information is uni-directional, only from lower- to higher-fidelity levels. As discussed in the previous section, as the number of levels of fidelity considered in the proposed architecture increases, it leads to an increase in the computational complexity, leading to longer training times. The lighter version provides an architecture with faster training, with some reduction in accuracy, owing to the reduction in network computations as the down-sampling operations are avoided. The choice between the proposed model and the lighter version can be made by the modeler based on the problem at hand, considering the trade-off between the accuracy and the training time. An overview of the  MF-UNet Lite architecture is given in Fig. \ref{fig:methodology_mf_b}. 

The illustration of the architecture in Fig. \ref{fig:methodology_mf_b} uses same three levels of fidelity, without loss of generality.  The training of the model starts at the lowest fidelity level, LR graph, with node attributes, $\bm{n}_3$, which are encoded and are, then, passed through the GN blocks to obtain the updated node attributes. These are, then, passed through the decoder for getting the predicted response at the low-fidelity level, $\hat{\bm{u}}_3$. Before the updated node attributes are passed to the decoder, they are also passed to the next level of fidelity, MR Graph, as shown in Fig. \ref{fig:methodology_mf_b}. The node attributes, $\bm{n}_2$, are processed initially, in a similar manner, for MR graphs as that of LR graphs, by passing through the encoder to get the encoded node attributes. But these are only passed through first few GN blocks, after which, they are added to the the updated node attributes from the LR graphs. The rest of the GN blocks process these coupled node attributes to obtain the updated node attributes for MR graphs. These are, then, passed through the decoder to get the predicted response, $\hat{\bm{u}}_2$. The updated node attributes are also passed to the next higher fidelity graph, which is the HR graph, in this case. A similar processing of graph happen for the HR graph as that of MR graph, where the node attributes processed after a few GN blocks are coupled (added) to the updated node attributes from the MR graph and passed through the rest of the GN blocks and then, through the decoder to get the predicted response, $\hat{\bm{u}}_1$.  We use the up-sampling process, based on $k$-nearest neighbors, described in the previous section, to pass the nodal information from lower to higher levels of fidelity. That is, we use $k$-nearest nodes based on Euclidean distance between the nodes to map every node in the LR graph to $k$ nodes in the MR graphs and similarly, that in the MR graph to $k$ nodes in the HR graphs. The node attributes from the LR graph is added to the node attributes of the k-nearest nodes in the MR graph and so on. The impact of the coupling of  node attributes from the lower levels is controlled by the weight parameter, $\beta$, which is learned during the training of the model. Additionally, the encoder, GN blocks and the decoder of all the levels of fidelity have shared parameters during the training of the model. 

Similar to MF-UNET, this architecture uses the same number of low-, medium- and high-fidelity data for training, as these graphs are coupled and trained together. All the required data are generated by running finite element simulations at different mesh resolutions and carrying out data pre-processing to convert the meshes to graphs and to extract node and edge information needed for training the model. The model is trained by using the predicted responses from all the three levels of fidelity,  $\hat{\bm{u}}_1$, $\hat{\bm{u}}_2$ and $\hat{\bm{u}}_3$. The respective ground truth for these graphs, $\bm{u}_1$, $\bm{u}_2$ and $\bm{u}_3$ from the finite element simulations are used for calculating the loss function for training the model.

\subsection{Training of the proposed GNN models}
\label{training_method}

The encoder and decoder are MLP networks with a single hidden layer and ReLU activation function. GN blocks consist of node and edge update modules as detailed in Section \ref{GNN}. The training of the proposed GNN models is done by updating the parameters of the encoder and decoder networks and the GN blocks in the network architecture, so that the loss function is minimized.  The loss function for both  models (MF-UNet and MF-UNet Lite) consists of $n$ terms, where $n$ is the number of levels of fidelity considered in the architecture. Let $L_i$ be the loss term corresponding to level $i$, which is calculated by comparing the predicted response of the graph at level $i$ to that of the ground truth for the graph of the same fidelity obtained through the finite element simulation.  Thus the total loss function, $L$, for training the proposed models can be written as:

\begin{equation}
    \begin{aligned}
    \label{eq:loss}
    L = \Sigma_{i=1}^n \lambda_i L_i
    \end{aligned}
\end{equation}
where $\lambda_i$ is the weight of the individual loss component, $L_i$, which is a hyper-parameter tuned during the training of the model. For the MF-UNet and MF-UNet Lite architectures given in Fig. \ref{fig:methodology_mf_a} and Fig. \ref{fig:methodology_mf_b} respectively, the number of levels, $n=3$. Therefore the loss function, $L = \lambda_1 L_1 + \lambda_1 L_2 + \lambda_1 L_3$, where $L_1$, $L_2$ and $L_3$ are the loss values corresponding to HR, MR and LR graphs respectively.

\section{Numerical Examples}
\label{results}

This section details the experimental setup utilized to validate the effectiveness of the proposed multi-fidelity approaches in approximating the solutions to PDEs, along with a discussion of the obtained results. The first two examples are 2D static problems - prediction of displacement on a cantilever beam and evaluation of stress concentration in a 2D plate. The final example is a 3D CFD problem evaluating the pressure and stress on vehicle models. Training for all the experiments are performed using PyTorch on Tesla V100-SXM2-16GB GPU. Pytorch geometric is used to process the graph data and to train all the models. 

\subsection{Displacement in a 2D cantilever beam}
\label{example_cantilever}

\begin{figure}  
     \centering
     \includegraphics[width=0.9\textwidth]{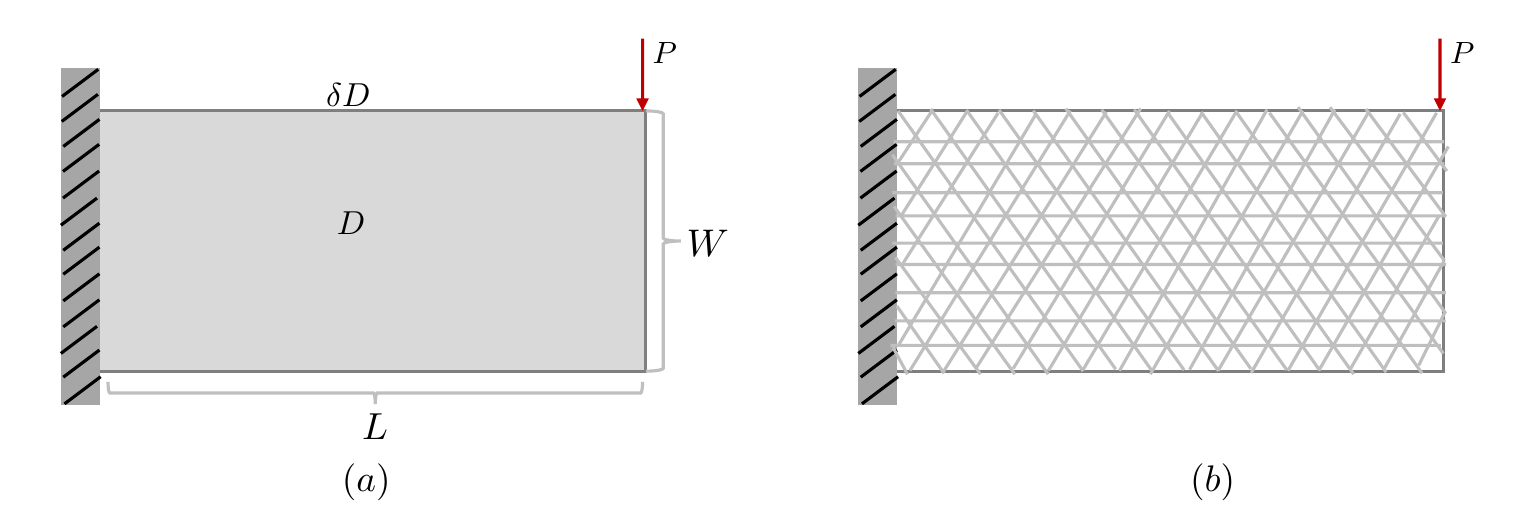}
     \caption{Cantilever beam with a fixed end considered for the multi-fidelity problem. (a) shows the domain along with the boundary, $\delta D$ as well as the force, $P$, applied as the boundary condition. (b) shows the triangular meshing of the domain.}
     \label{fig:cantilever}
\end{figure}

For the first example, we assess the performance of MF-UNet and MF-UNet Lite models in evaluating the displacement of a beam subject to an external loading. For this, we consider a cantilever beam in an elasto-static solid mechanics problem setup. This is a benchmark problem introduced in \cite{black2022learning}. The underlying governing equations for this problem can be written in the form of the following compatibility equations. 

\begin{equation}
    \begin{aligned}
    &\frac{\partial u_x}{\partial x} = \frac{1}{E}\left(\sigma_{xx}-\nu \sigma_{yy}\right), \\
    &\frac{\partial u_y}{\partial y} = \frac{1}{E}\left(\sigma_{yy}-\nu \sigma_{xx}\right) ,\\
    &\frac{1}{2} \left(\frac{\partial u_x}{\partial y} + \frac{\partial u_y}{\partial x}\right) = \frac{1}{G}\left(\sigma_{xy}\right), \\
\label{eq_compatibility}
    \end{aligned}
\end{equation}
where $E$ is the elastic modulus and $\nu$ is the Poisson's ratio and we solve the displacement field, $\bm{u}$, given the boundary conditions consisting of Dirichlet and Neumann BCs and body forces on a generalized domain, $\mathcal{D}$. We consider a cantilever beam with length $L$ and height $H$ with a fixed end and an elastic material with the Elastic modulus of $E = 200$ GPa and Poisson’s ratio of $\nu = 0.3 $ under isotropic, 2D, linear-elastic plane-strain conditions. The length and height of the beam are randomly selected in the training data. The fixed end can be either in the left or right edge of the beam. It is subjected to a force with a fixed total magnitude. The direction, location of the force is randomly selected.  The applied force can be either a single nodal force with the magnitude $10^6$ N or a uniformly distribution load with the total  magnitude of $10^6$ N. Furthermore,  $L$ varies between $6$ and $15$ m and $H$ varies between $3$ and $9$ m. We use a triangular mesh with for the domain with a random uniform noise limited by $0.1$ applied to the mesh discretization by adding it to the nodal coordinates. We carried out FEM simulations in FeNics \cite{alnaes2015fenics} to generate 3,000 low-, medium- and high-fidelity meshes and the corresponding solutions. Fig. \ref{fig:cantilever} shows the domain of the cantilever beam along with the meshing of the domain.  The meshes are converted to graphs (low-, medium- and high-resolution) and node attributes are generated for the samples. The node attributes considered for the model are nodal coordinates ($x$ and $y$), location and direction of the force and type of node (boundary or interior). The edge attributes are the relative distance in $x$ and $y$ directions. The response evaluated for training is the displacement in $x$ and $y$ directions, $u_x$ and $u_y$. Low-, medium- and high-resolution graphs have an average of 200, 400 and 700 nodes respectively.

The efficacy of the proposed GNN architectures, MF-UNet and MF-UNet Lite, is evaluated against two benchmark models: 1) a single-fidelity GNN, which is trained using only high-fidelity samples, and 2) transfer learning GNN model \cite{liu2022multi}, which uses low-fidelity data to train the GNN model initially, which is used as the pre-trained network to train using the high-fidelity data. Since, MF-UNet and MF-UNet Lite and transfer learning use graphs of different resolutions, single-fidelity GNN training is not conducted with the same number of high-resolution graphs. We adjust the size of the training dataset for all the models, by equalizing the computational time required for generating the respective datasets needed for training each of the models. We train MF-UNet and MF-UNet Lite and transfer learning using two and three levels of fidelity to compare the impact of number of fidelities considered. The models trained using two levels of fidelity are referred to as MF-UNet-2, MF-UNet Lite-2 and Transfer Learning-2 respectively. Similarly, for three levels of fidelity, they are referred to as MF-UNet-3, MF-UNet Lite-3 and Transfer Learning-3 respectively. For models with two levels of fidelity, we use 2,000 low-resolution and high-resolution graphs for training, whereas for those with three levels of fidelity, we use 1,500 low-, medium- and high-resolution graphs. For single-fidelity model, we use 2,700 high-resolution graphs. 

For the GNN architecture for the benchmark and proposed models, we have an encoder, a series of GN blocks and a decoder. The encoder is an MLP network of single hidden layer with architecture 11-64-128 (i.e. 11 inputs nodes and 64 hidden layer nodes and 128 output nodes). There is another encoder, with similar architecture 3-64-128, that encodes the edge attributes, which are the relative distances in $x$ and $y$ directions as well as the Euclidean distance between the nodes. The decoder is an MLP network of single hidden layer with architecture 128-64-2. ReLU is used as the activation function for both the encoders and the decoder. The GN block consists of node and edge update modules, as detailed in Section \ref{GNN}. We use 10 GN blocks for single-fidelity and transfer learning models, with ReLU as the activation function for the node and edge update networks. For MF-UNet and MF-UNet Lite, the low-resolution (LR) graphs use 10 GN blocks for message passing and aggregation. For medium- and high-resolution graphs, flow (coupling) of information from and to the other levels are done after 5 GN blocks, as shown in Fig. \ref{fig:methodology_mf_a} and Fig. \ref{fig:methodology_mf_b} respectively. 

We use the $k$-nearest neighbors algorithm to map nodes between graphs of different fidelity levels for the transfer of node attributes. For this problem, we use $k=4$. The $k$-nearest neighbors for all the graphs in the dataset are calculated offline during the data generation step to avoid additional computation time during the training process. Once the node attribute information from the lower level of fidelity is added to the node attributes to the current level, the graph is passed through another 5 GN blocks to get the updated node attributes. Thus, MF-UNet and MF-UNet Lite also have 10 GN blocks in total, like single-fidelity and transfer learning models. The same decoder architecture as other models is used for evaluating the nodal stress concentration from all the three fidelity graphs. For training all the models, we use Adam optimization with a learning rate of $0.0002$ and a weight decay of $1E-6$. We use cosine annealing with warm restart as the learning rate scheduler, and the mean squared error (MSE) as the loss function. For MF-UNet and MF-UNet Lite, we use the loss function defined in Eq.~\eqref{eq:loss}. Here, the individual loss terms would be mean squared error calculated using the predicted response and ground truth at each level of fidelity. For MF-UNet-2 and MF-UNet Lite-2, we use $\lambda_1=10$ and  $\lambda_2=1$, whereas for MF-UNet-3 and MF-UNet Lite-3, we use $\lambda_1=10$, $\lambda_1=5$ and  $\lambda_3=1$. All the models are trained for 2,000 epochs, with a batch size of 1. In order to compare the prediction power of the  models, we use relative L1-error, defined as

\begin{equation}
    \begin{aligned}
    e(\bm u) = \dfrac{ \|\widehat{\bm u} - \bm u\|_1}{\|\bm u\|_1},
\label{relative_error}
    \end{aligned}
\end{equation}
where $\bm u$ denotes the ground truth nodal displacements and $\widehat{\bm u}$ is the GNN prediction of nodal displacements, and $\| \cdot \|_1$ denotes the $\ell_1$ norm. 

\begin{table}[!ht]
\begin{center}
\caption{The number of model parameters and relative L1-error in predicting nodal displacements, $u_x$ and $u_y$, for the high-resolution graphs in the testing dataset for the cantilever beam problem for the benchmark GNN models and the proposed multi-fidelity GNN approaches.}
\begin{tabular}{lccr}
\toprule
Model & \# Parameters & $e(u_x)$ & $e(u_y)$  \\
\midrule
Single Fidelity & 1,705,281 & 73\% & 62\% \\
Transfer Learning-2 & 1,705,281 & 68\% & 52\% \\
Transfer Learning-3 & 1,705,281 &  46\% & 30\% \\
MF-UNet-2 & 1,705,281 & 10\% & 3\% \\
MF-UNet-3 & 1,705,281 & \textbf{8\%} & \textbf{2\%} \\
MF-UNet Lite-2 & 1,705,281 & 13\% & 6\% \\
MF-UNet Lite-3 & 1,705,281 & 11\% & 5\% \\
\bottomrule
\end{tabular}
\label{table:3}
\end{center}
\end{table}

\begin{figure}  
     \centering
     \includegraphics[width=0.8\textwidth]{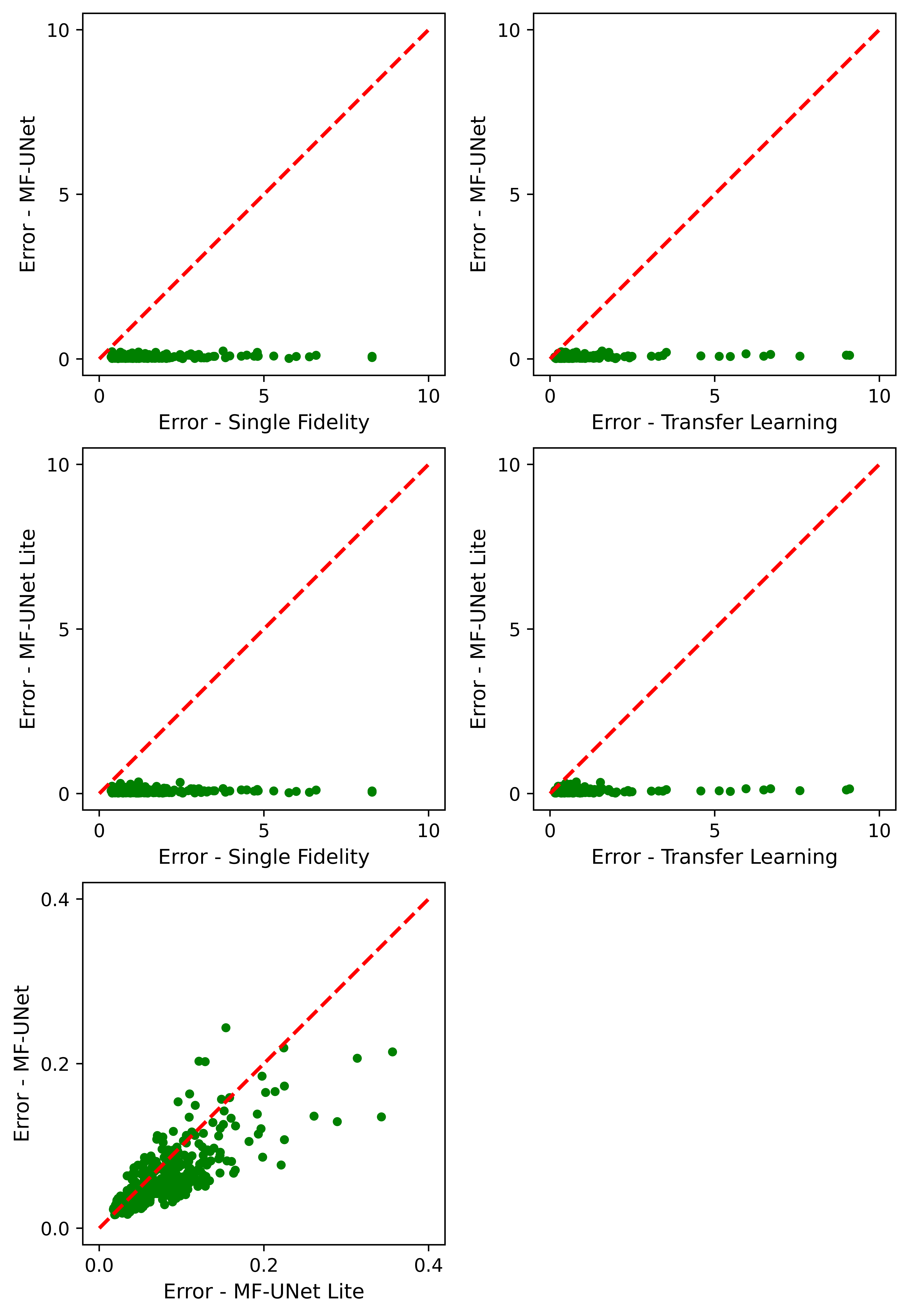}
     \caption{A comparison of relative L1-error in the prediction of $u_x$ for individual high-resolution test samples between the benchmark GNN models and the proposed multi-fidelity models, for the cantilever beam problem. Here, 3 levels of fidelity are considered for training Transfer Learning, MF-UNet and MF-UNet Lite.}
     \label{fig:canti_error_all_models}
\end{figure}

Table \ref{table:3} shows the relative L1-error for the prediction of $u_x$ and $u_y$ for high-resolution graphs for different models on the testing dataset along with the number of model parameters. From these values, we can observe that the proposed models, MF-UNet and MF-UNet Lite, with 2 and 3 levels of fidelity, perform significantly better than the benchmark GNN models. Single fidelity GNN is the worst performing model. As the level of fidelity increases, the transfer learning model improves in performance, but falls short significantly compared to the proposed models. The given sample size used for training seems to be insufficient for the transfer learning approach to learn the responses of the physical system accurately. This performance could be improved by training with a much larger sample of low-fidelity data. MF-UNet with three levels of fidelity is the best performing model, with 1-3\% improvement in the relative L1-error from that with two levels of fidelity. MF-UNet Lite has around 3\% increase in relative L1-error compared to that of MF-UNet for both two and three levels of fidelity. Fig. \ref{fig:canti_error_all_models} shows comparison plots of the relative L1-error for predicting $u_x$ for all the high-resolution graphs in the testing dataset for different pairs of benchmark and proposed models. They highlight the significant improvement in the prediction capability of the proposed models over the benchmark GNN models. All the models have same number of parameters, since the entire GNN architecture (encoder, GN blocks and decoder) are shared between different levels of fidelity. Thus, adding different levels of fidelity does not increase the model complexity of the proposed models. 

From the above analyses, it is evident that using multi-fidelity data and architecture has a significant impact on the performance of the models.  It is to be noted that, the only difference between the network architecture of transfer learning to that of MF-UNet is the addition of coupling of node attributes across different levels of fidelity. Thus, the improved performance of MF-UNet over transfer learning can be attributed to this. 



\begin{figure}
     \begin{subfigure}[b]{0.49\textwidth}
         \centering
         \includegraphics[width=0.8\textwidth]{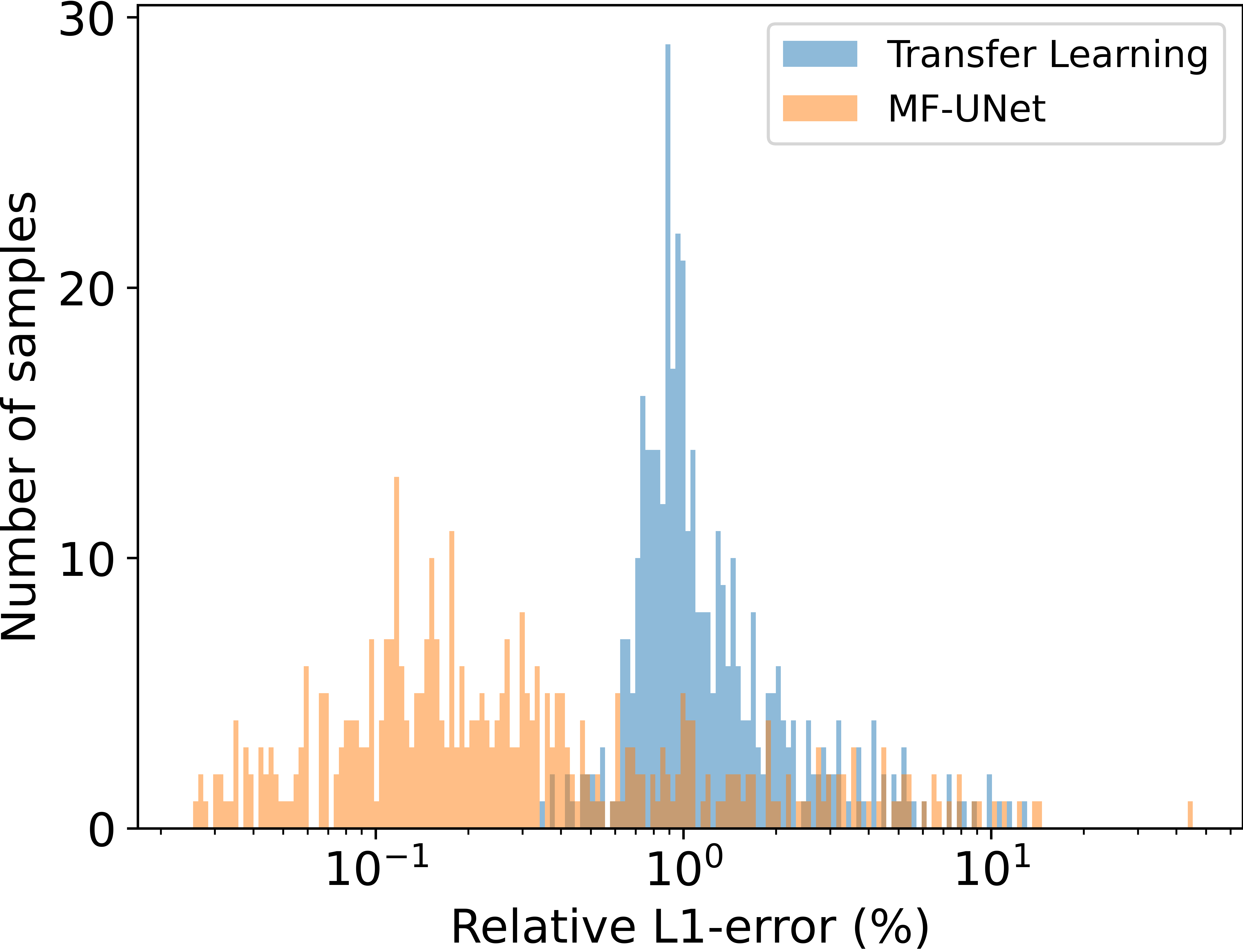}
         \caption{$\hat{u}_x$}
         \label{fig:mf_u_hist}
     \end{subfigure}
    \hfill
     \begin{subfigure}[b]{0.49\textwidth}
         \centering
         \includegraphics[width=0.8\textwidth]{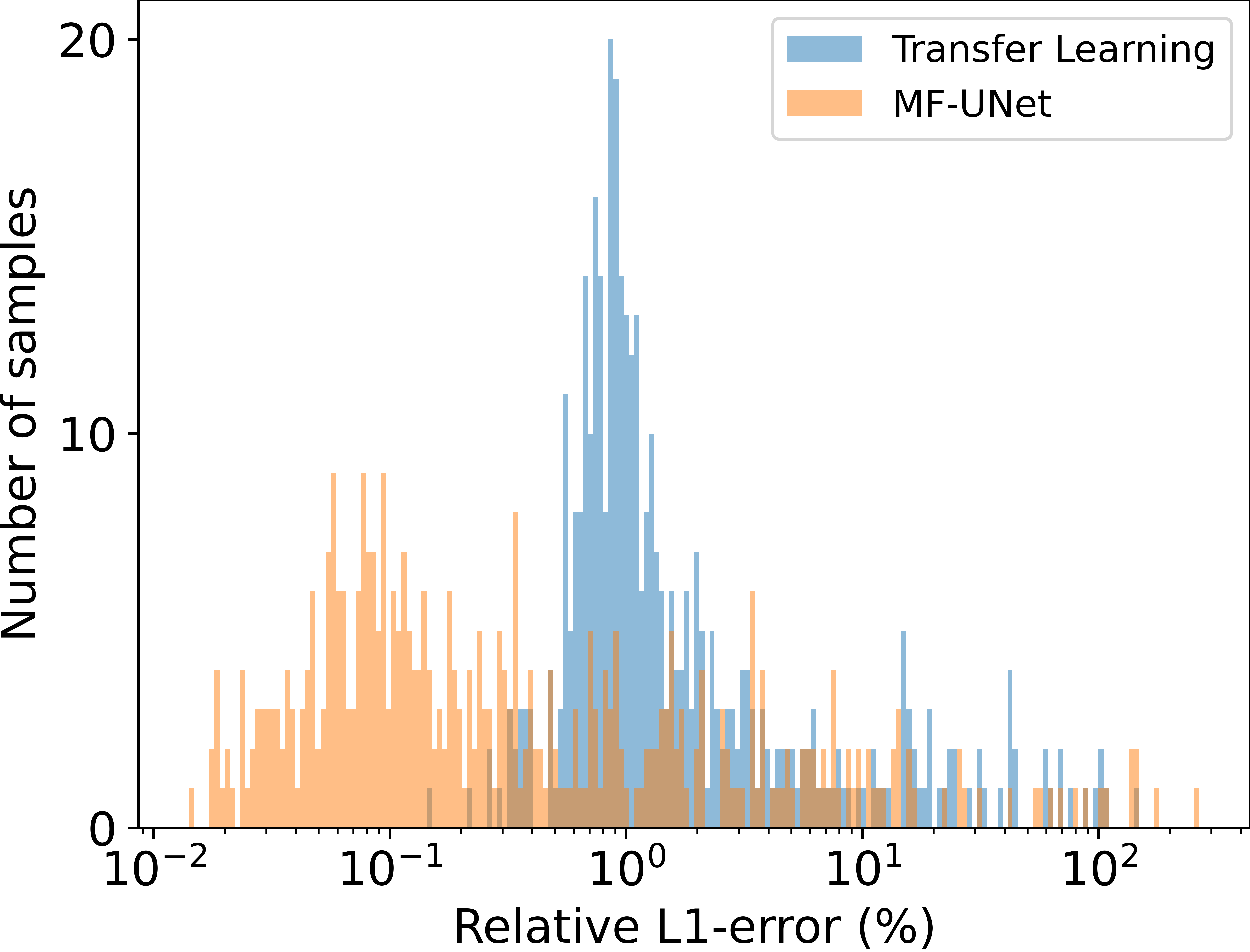}
         \caption{$\hat{u}_y$}
         \label{fig:mf_v_hist}
     \end{subfigure}

     \caption{A histogram showing the distribution of the relative L1-error in the prediction of displacement for graphs whose resolutions are 3-times that of the high-fidelity graphs used for training transfer learning GNN and MF-UNet model. This shows the generalization capability of MF-UNet to predict the responses on graphs with unseen resolutions.}
    \label{fig:mf_hist}
\end{figure}

We also evaluate the generalization capability of the MF-UNet architecture to predict the responses of graphs with resolutions unseen during the training. For this, we consider graphs with resolution 3-times that of the highest fidelity used for training MF-UNet and its performance is compared against the transfer learning GNN, since it is the best performing benchmark model. We consider 3 levels of fidelity for both the models. Fig. \ref{fig:mf_hist} presents  the performance comparison by showing histograms of the relative L1-error in displacement prediction for $u_x$ and $u_y$ for a sample of graphs. This highlights the significant improvement in the generalization capability to 3$\times$ finer graphs for MF-UNet over the transfer learning method. 

\subsubsection{Ablation study}

\begin{figure}
     \begin{subfigure}[b]{0.49\textwidth}
         \centering
         \includegraphics[width=0.8\textwidth]{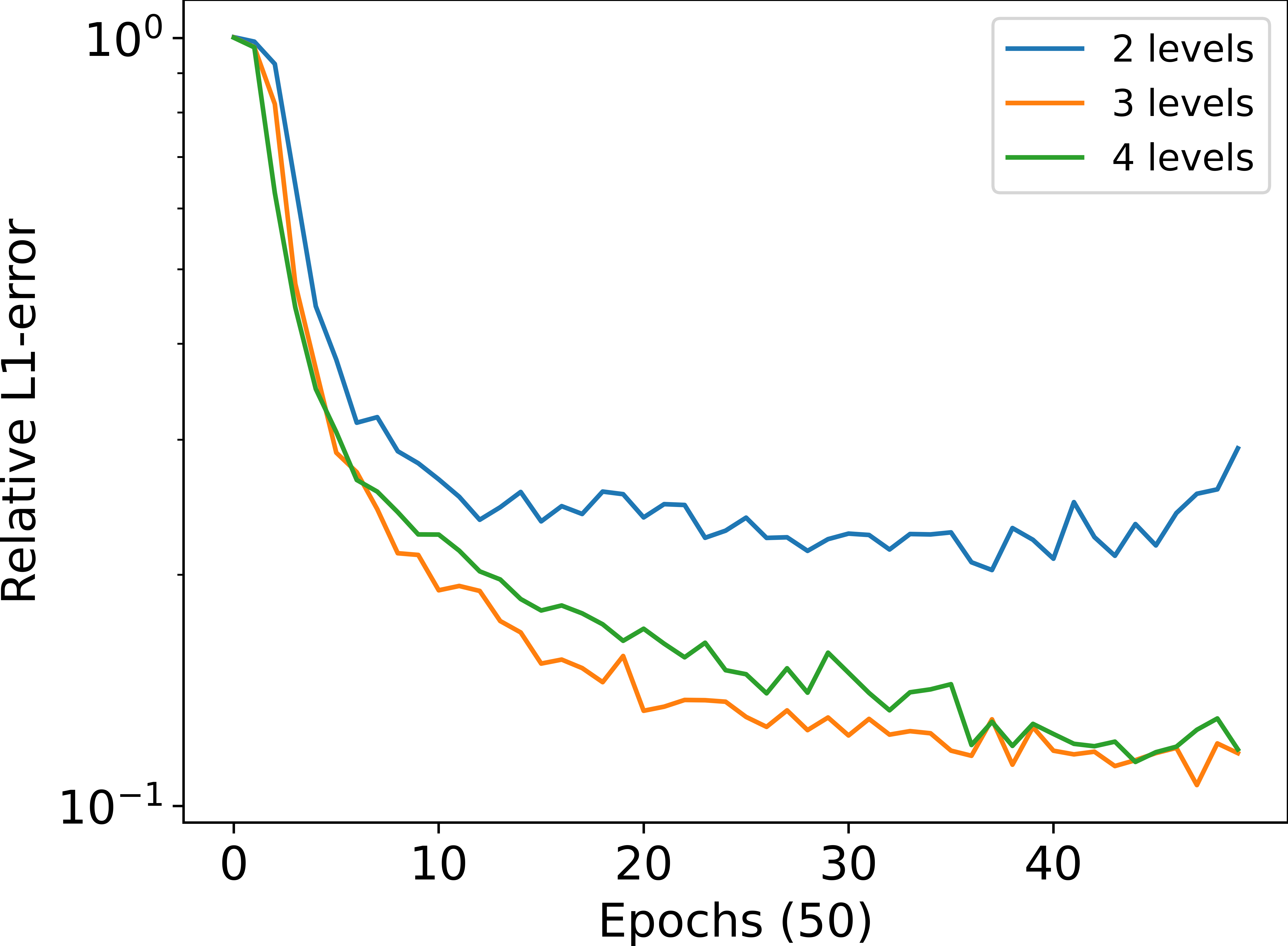}
         \caption{$\hat{u}_x$}
         \label{fig:mf_u_ablation_levels}
     \end{subfigure}
    \hfill
     \begin{subfigure}[b]{0.49\textwidth}
         \centering
         \includegraphics[width=0.8\textwidth]{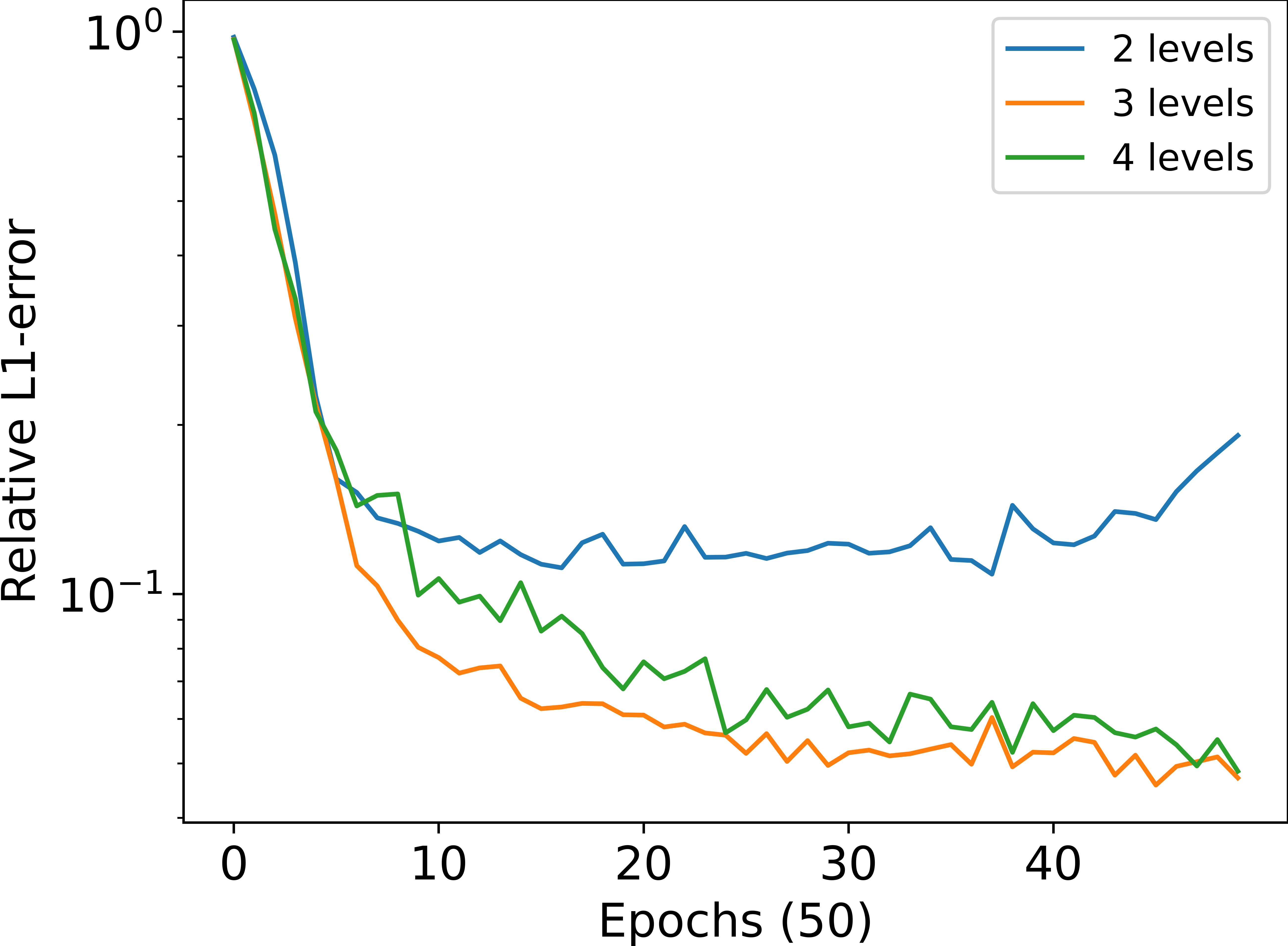}
         \caption{$\hat{u}_y$}
         \label{fig:mf_v_ablation_levels}
     \end{subfigure}

     \caption{An analysis on the impact of the choice of the number of fidelity or resolution levels used in MF-UNet architecture on the relative L1-error for the prediction of $\hat{u}_x$ and $\hat{u}_y$.  The  errors are calculated using predictions on the high-resolution nodes.}
    \label{fig:ablation_levels}
\end{figure}

We carry out ablation study to understand the impact of the number of levels (resolutions) and the distance between these resolutions on the performance of MF-UNet model. For the former, we train MF-UNet with 2, 3 and 4 levels of resolutions, after equalizing the training data size for fair comparison. Fig. \ref{fig:ablation_levels} shows how the relative L1-error for the prediction of $u_x$ and $u_y$ decrease over the epochs for different MF-UNet models. While we see a decrease in error from 2 to 3 levels, we do not see a significant change going beyond 3 levels.

\begin{figure}
     \begin{subfigure}[b]{0.49\textwidth}
         \centering
         \includegraphics[width=0.8\textwidth]{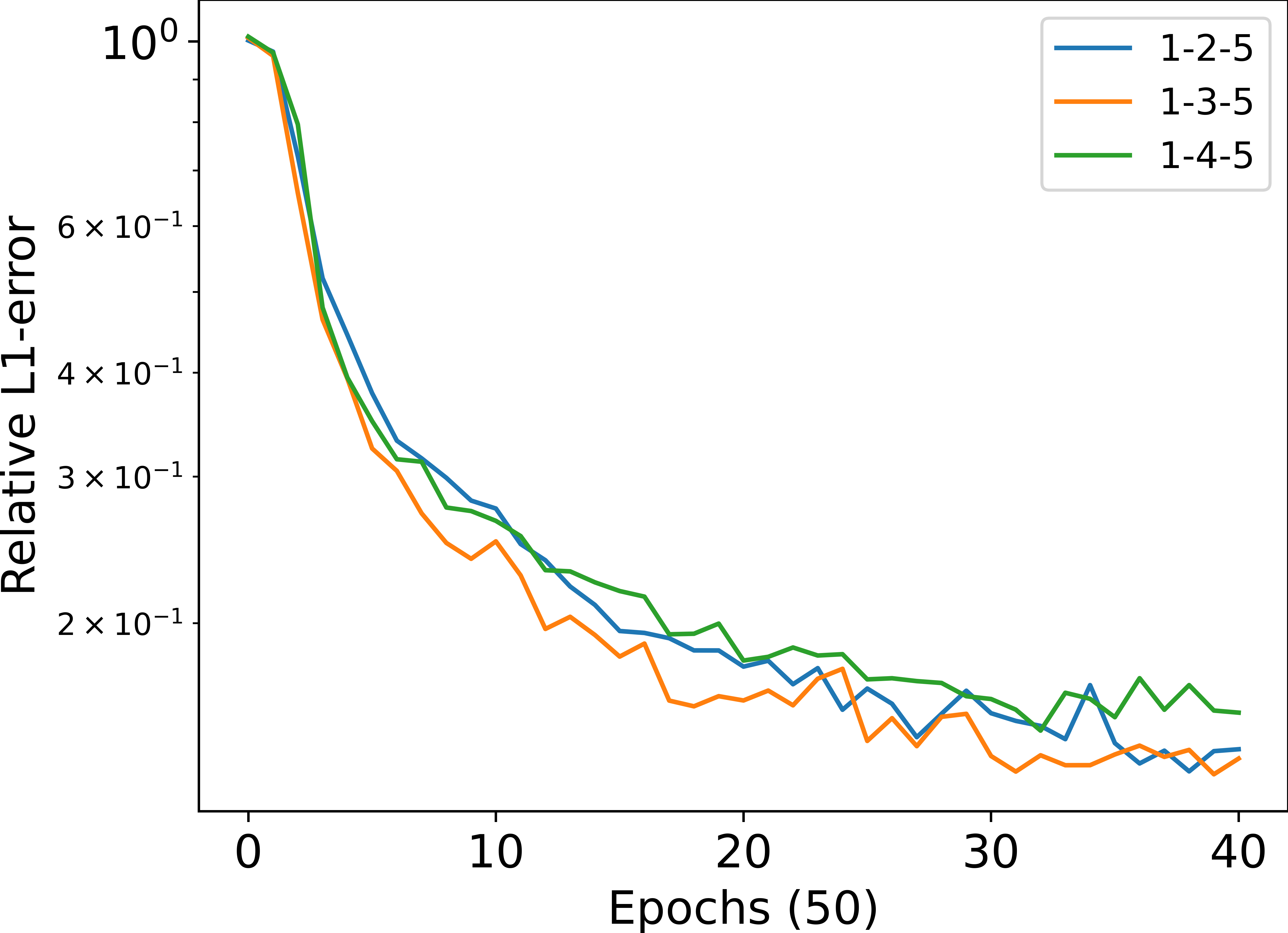}
         \caption{$\hat{u}_x$}
         \label{fig:mf_u_ablation_resol}
     \end{subfigure}
    \hfill
     \begin{subfigure}[b]{0.49\textwidth}
         \centering
         \includegraphics[width=0.8\textwidth]{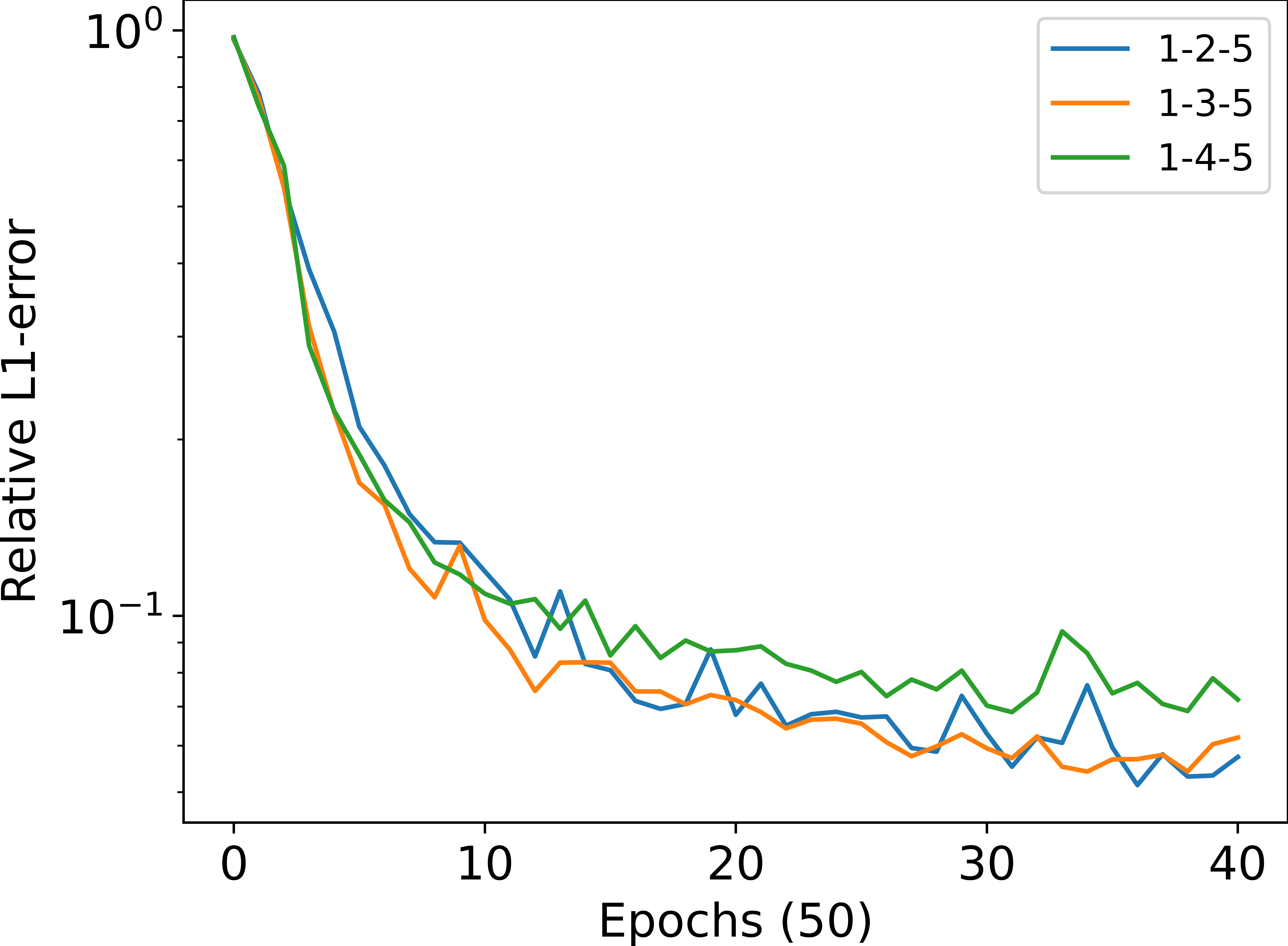}
         \caption{$\hat{u}_y$}
         \label{fig:mf_v_ablation_resol}
     \end{subfigure}

     \caption{An analysis on the impact of the choice of resolution at different levels of MF-UNet architecture on the L1-relative error for the prediction of $\hat{u}_x$ and $\hat{u}_y$. 1-2-5 indicates that the low-, medium- and high-resolution graphs are of 1x, 2x and 5x resolutions, i.e., the high-resolution graphs have five times the number of nodes compared to that of low-resolution graphs. Similarly, 1-3-5 and 1-4-5 indicates graphs with 1x, 3x and 5x resolutions and 1x, 4x and 5x resolutions, respectively. The  errors are calculated using predictions on the high-resolution nodes.}
    \label{fig:ablation_135}
\end{figure}

Furthermore, to understand the impact of the choice of resolution on the performance, we train MF-UNet with three levels, but the ratios between the resolutions are varied. In particular, we consider three cases: (a) 1-2-5, where medium- and high-resolution graphs have 2 and 5 times the number of nodes compared to low-resolution graphs, respectively; (b) 1-3-5, which is similar to the first case except for the medium-resolution graph which now has 3 times the number of nodes as that of low-resolution graphs; and (c) 1-4-5, where the medium-resolution graphs have 4 times the number of nodes as that of low-resolution, while the high-resolution graph remains the same. Fig. \ref{fig:ablation_135} shows the impact of the change in resolutions on the performance of MF-UNet. We do not observe a significant change in error values between the three cases, thus concluding that changing ratios between the resolutions in different levels does not affect the performance of MF-UNet.

\subsection{Stress distribution in a 2D plate}
\label{2dplate_results}

    In the second example, we assess the effectiveness of the proposed multi-fidelity GNN architectures to evaluate the stress concentration in 2D rectangular steel plates. In particular, we consider two cases introduced in \cite{taghizadeh2024multifidelity}: (1) rectangular steel plates with notches at the top and bottom edges and (2) steel plates with an interior hole whose position varies within the plate boundaries. For both cases, we set the Young's Modulus $E=210$ GPa, the density $\rho=7800 \text{kg/m3}$ and Poisson's ratio, $\nu = 0.3$. The length ($L$), width ($W$) and thickness ($T$) of the plates are set at 0.4 m, 0.1 m and 0.001 m, respectively. The horizontal movement of the plate is constrained on the left side and the vertical movement is restricted at the mid-point of the left side of the plate. For the notched rectangular plate, the ratio of the notch diameter to the plate width is varied between $0.2$ and $0.5$ , and for the plate with the hole, the ratio of the diameter of the hole to the plate width is varied between $0.2$ and $0.5$. For both  cases, the applied load ranges from $1000 \text{N}$ to $2000 \text{N}$. Both the datasets are generated using the PyAnsys \cite{alexander_kaszynski_2020_4009467} software package. 


\begin{figure}
     \begin{subfigure}[b]{0.49\textwidth}
         \centering
         \includegraphics[width=\textwidth]{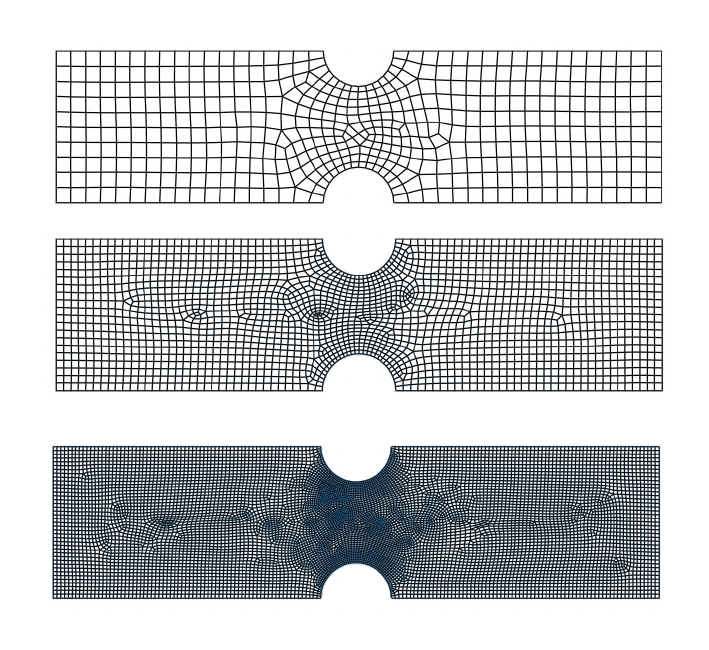}
         \caption{Plate with notches}
         \label{fig:sample_notches}
     \end{subfigure}
    \hfill
     \begin{subfigure}[b]{0.49\textwidth}
         \centering
         \includegraphics[width=\textwidth]{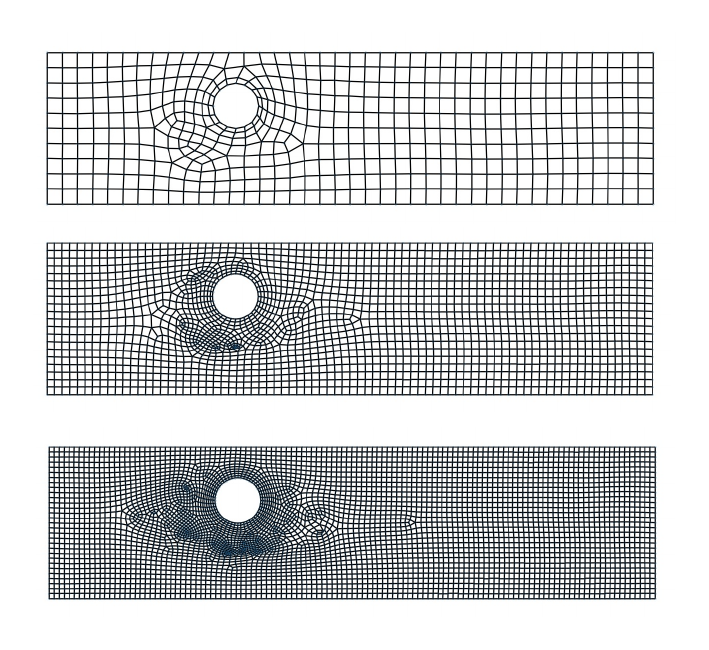}
         \caption{Plate with variable hole}
         \label{fig:sample_variable}
     \end{subfigure}

     \caption{A sample training data generated for 2D plates with (a) variable notches  and (b) variable hole. For each case, three resolutions of the graphs (meshes) are shown.}
    \label{fig:2dplate_sample_resolution}
\end{figure}

\begin{figure}
     \begin{subfigure}[b]{0.49\textwidth}
         \centering
         \includegraphics[width=\textwidth]{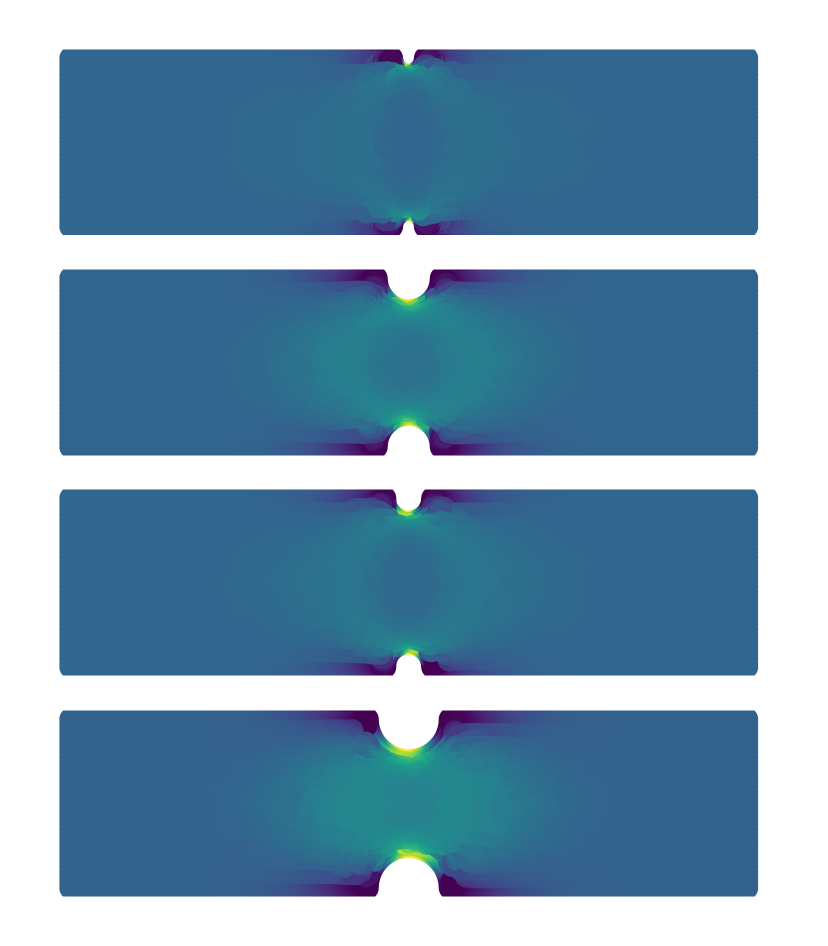}
         \caption{Plate with notches}
         \label{fig:stress_notches}
     \end{subfigure}
    \hfill
     \begin{subfigure}[b]{0.49\textwidth}
         \centering
         \includegraphics[width=\textwidth]{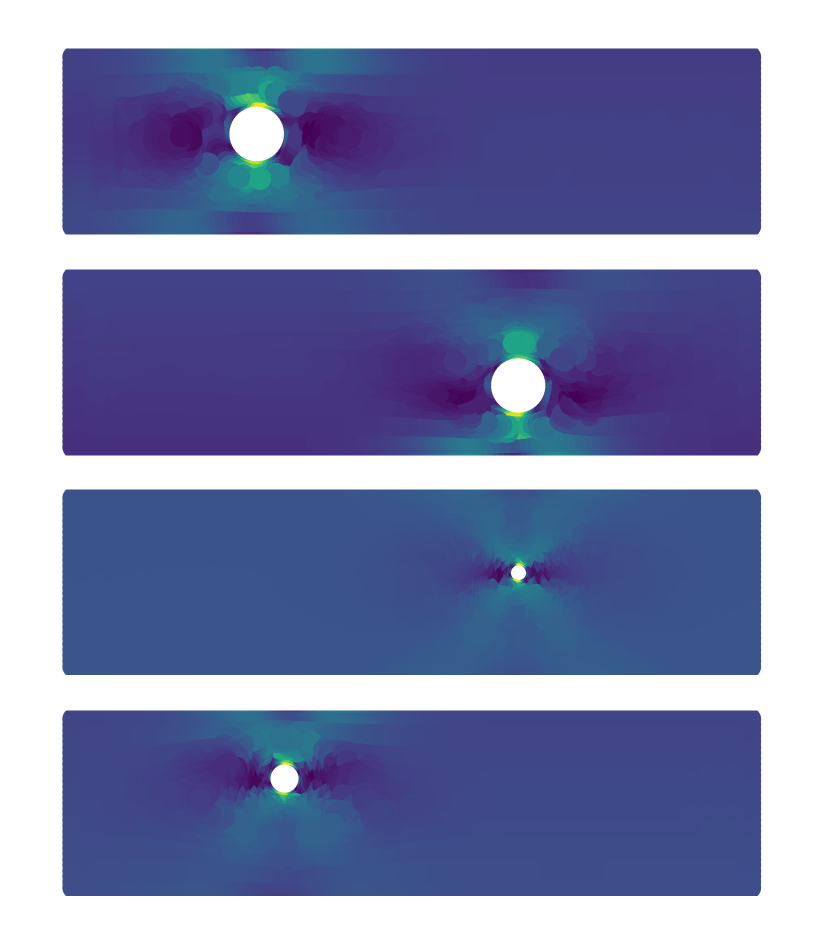}
         \caption{Plate with variable hole}
         \label{fig:stress_variable_hole}
     \end{subfigure}

     \caption{Three samples of training data generated for 2D plates with (a) variable notches and (b) variable hole. Also shown is the distribution of von Mises stress calculated using the finite element method.}
    \label{fig:2dplate_sample_stress}
\end{figure}

The simulation training datasets are obtained at three levels of fidelity (or mesh resolutions) for both the notched and variable hole experiments. The low-fidelity data consists of coarse meshes with an average of 580 nodes, medium-fidelity data has meshes with an average of 5,200 nodes, whereas high-fidelity data uses fine meshes, with an average of 9,821 nodes. Fig.~\ref{fig:2dplate_sample_resolution} shows different mesh resolutions for a representative training sample for each of these two cases. Furthermore, Fig.~\ref{fig:2dplate_sample_stress} demonstrate  the variation in geometry on three different training samples for these two cases along with the distribution of von Mises stress.

The performance of the proposed GNN architectures, i.e., MF-UNet and MF-UNet Lite, is evaluated against the benchmark models, namely the single-fidelity and transfer learning GNNs, which are described in the previous numerical experiment. Similarly to the previous example, we compare the performance of the proposed models for 2 and 3 levels of fidelity, i.e. via MF-UNet-2 and MF-UNet-3 as well as MF-UNet Lite-2 and MF-UNet Lite-3. Therefore, we consider the same levels of fidelity for the benchmark transfer learning model as well, referring to them as Transfer Learning-2 and Transfer Learning-3. We adjust the size of the training dataset for all the models, by equalizing the computational time required for generating the respective datasets needed for training  each of the models. For models with 2 levels of fidelity, we use 909 low-resolution and high-resolution graphs for training, whereas for those with 3 levels of fidelity, we use 625 low-, medium- and high-resolution graphs. For single-fidelity model, we use 1,000 high-resolution graphs. The additional high-resolution graphs used for training the single-fidelity model equalize the computational time needed to generate 909 low-resolution graphs for Transfer Learning-2, MF-UNet-2, and MF-UNet Lite-2 as well as 625 low- and medium-resolution graphs for Transfer Learning-3, MF-UNet-3, and MF-UNet Lite-3. 

The GNN architecture for all the models consists of an encoder, a series of GN blocks and a decoder. The encoder is an MLP network of single hidden layer with the architecture 10-64-128 (i.e. 10 inputs nodes and 64 hidden layer nodes and 128 output nodes). The encoder network encodes the node attributes, which are chosen to include  nodal coordinates, diameters of the notches and holes, material properties, and loading and boundary conditions. There is another encoder, with similar architecture 3-64-128, that encodes the edge attributes, which are the relative distances in $x$ and $y$ directions as well as the Euclidean distance between the nodes. The decoder is an MLP network of single hidden layer with architecture 128-64-1. The output is the von Mises stress at every node. ReLU is used as the activation function for both the encoders and the decoder. The GN block consists of node and edge update modules, as detailed in Section \ref{GNN}. Here, after a number of trials,  we chose 10 GN blocks for all the models, with ReLU as the activation function for the node and edge update networks. For MF-UNet and MF-UNet Lite, the flow of nodal information between different levels of fidelity (as shown in Fig. \ref{fig:methodology_mf_a} and Fig. \ref{fig:methodology_mf_b} respectively) is done after the fifth GN block. We use $k=3$ for finding $k$-nearest neighbors to map the nodes between graphs of different resolutions. Similar to the previous example, the $k$-nearest neighbors of all the graphs are calculated offline during the dataset generation step. For training all the models, we use Adam as the optimizer with a learning rate of $0.001$ and a weight decay of $1\times10^-6$. We use cosine annealing with warm restart \cite{loshchilov2017fixing} as the learning rate scheduler. For training, we use the relative L2-error for all the nodes across the geometry samples in a batch size as the loss function, which is defined as 

\begin{equation}
    \begin{aligned}
    e_{\ell_2} = \dfrac{ \|\widehat{\bm \sigma} - \bm \sigma \|_2}{\|\bm \sigma\|_2},
\label{relative_error_l2}
    \end{aligned}
\end{equation}
where $\bm \sigma$ is the exact von Mises stress, and $\widehat{\bm \sigma}$ is the GNN prediction of this stress. $\| \cdot \|_2$ denotes the $\ell_2$ norm. For MF-UNet and MF-UNet Lite, $e_{\ell_2}$ is used for calculating the individual loss component for each fidelity as defined in Eq.~\eqref{eq:loss}. Similarly to the first example, for MF-UNet-2 and MF-UNet Lite-2, we use $\lambda_1=10$ and  $\lambda_2=1$, whereas for MF-UNet-3 and MF-UNet Lite-3, we use $\lambda_1=10$, $\lambda_2=5$ and  $\lambda_3=1$. All the models are trained for 2,000 epochs with a batch size of 2. 

\begin{table}[!ht]
\begin{center}
\caption{The number of model parameters and the relative L2-error as defined in Eq.~\eqref{relative_error_l2} across the testing dataset for different GNN models for evaluating the von Mises stress distribution  for 2D plates with variable notches and holes. }
\begin{tabular}{lccr}
\toprule
Model & \# Parameters & Plate with notches & Plate with a hole \\
\midrule
Single Fidelity & 1,705,281 & 8.1\% & 15.2\% \\
Transfer Learning-2 & 1,705,281 & 0.26\% & 5.1\% \\
Transfer Learning-3 & 1,705,281 &  0.21\% & 3.7\% \\
MF-UNet-2 & 1,705,281 & 0.10\% & 0.7\% \\
MF-UNet-3 & 1,705,281 & \textbf{0.09\%} & \textbf{0.3\%} \\
MF-UNet Lite-2 & 1,705,281 & 0.13\% & 1.5\% \\
MF-UNet Lite-3 & 1,705,281 & 0.11\% & 1.2\% \\
\bottomrule
\end{tabular}
\label{table:2dplate_errors}
\end{center}
\end{table}

Table \ref{table:2dplate_errors} presents a comparison of performance of different GNN models for both of  cases in terms of the relative L2-error calculated on  the test dataset. For both the variable notches and variable hole cases, the performance of the models is similar to the previous experiment, with all the variants of proposed models, MF-UNet and MF-UNet Lite, consistently outperforming the single-fidelity GNN model. The performance of the transfer learning approach for the notched plates is comparable to that of the proposed models. However, in the variable hole case, the performance disparity is more substantial, with MF-UNet and MF-UNet Lite models showing significantly better results. This is due to the higher complexity of the variable hole case, which requires more training data for capturing this more complex behavior. Our proposed models however can handle this complex case more effectively by also learning the coupling of behaviors across different fidelity levels. Overall, MF-UNet is consistently the best performing model for both  cases, followed by its lighter version, MF-UNet Lite. All the models have similar model complexity, with the same number of model parameters, as the encoder, GN blocks and the decoder are shared across different levels of fidelity. It is to be noted that the training time for MF-UNet is approximately 1.5 times than that of the single-fidelity model for the 2-level case and 1.8 times for the 3-levels case, owing to the processing of multiple graphs of different resolutions in a single network during training for the proposed architectures. However, this computational impact is offset by around 2 orders of magnitude improvement in the accuracy of the proposed models, with no addition in the model complexity. 


\subsubsection{Impact of bi-directional flow of information}

For this experiment, we also analyze the impact of bi-directional flow of information from high- to low-fidelity levels and vice verse as used in MF-UNet against the uni-directional flow from low- to high-fidelity levels in MF-UNet Lite. We use only two levels for both MF-UNet and MF-UNet Lite, and ensure that the only difference between the models is the uni- vs bi-directional flow of information between the resolutions. We use the plate with holes dataset for training both the models with varying number of GN blocks. Table. \ref{table:gnblock_variable} shows the relative L2-error for predicting the nodal von Mises stress for the testing data samples for MF-UNet Lite and MF-UNet. We can observe that for shallower networks with low number of GN blocks, the impact of bi-directional flow is significant with around $3\%$ difference in the relative L2-error, and as more GN blocks are added, the performance disparity between the models diminishes. Thus, making the flow of nodal information bi-directional can help reduce the model complexity of the model by making the network shallower, compared to a network with uni-directional flow with similar performance. However, MF-UNet Lite is  35\% faster in terms of training time per iteration, compared to MF-UNet. This can be attributed to the bi-directional flow of information in MF-UNet increasing the computations during training. Thus, there is a trade-off between the accuracy and the training time, which needs to be considered by the modeler when choosing the right model for the problem at hand.

We also studied the impact of the location of the coupling, which is the point of the network where the nodal information is transferred. We compared the performance of the models when the information is transferred at the beginning, middle and the end of the series of GN blocks at each level. However, we did not observe any significant change in model performance with the location of the coupling. 

\begin{table}[!ht]
\begin{center}
\caption{The relative L2-error as defined in Eq.~\eqref{relative_error_l2} across the testing dataset  for evaluating the von Mises stress concentration for 2D plates with holes for MF-UNet Lite and MF-UNet with two levels of fidelity and varying number of GN blocks. }
\begin{tabular}{lcr}
\toprule
GN blocks & MF-UNet Lite & MF-UNet \\
\midrule
2 & 12.2\%  & 9.5\%  \\
4 & 8.3\%  & 6.6\%  \\
6 & 5.0\% &  4.0\%  \\
8 & 3.7 \% & 2.8\%  \\
10 & 1.5\%  & 0.7\%  \\
\bottomrule
\end{tabular}
\label{table:gnblock_variable}
\end{center}
\end{table}

\subsection{Vehicle aerodynamics simulation for 3D geometries}

\begin{figure}
     \begin{subfigure}[b]{0.49\textwidth}
         \centering
         \includegraphics[width=\textwidth]{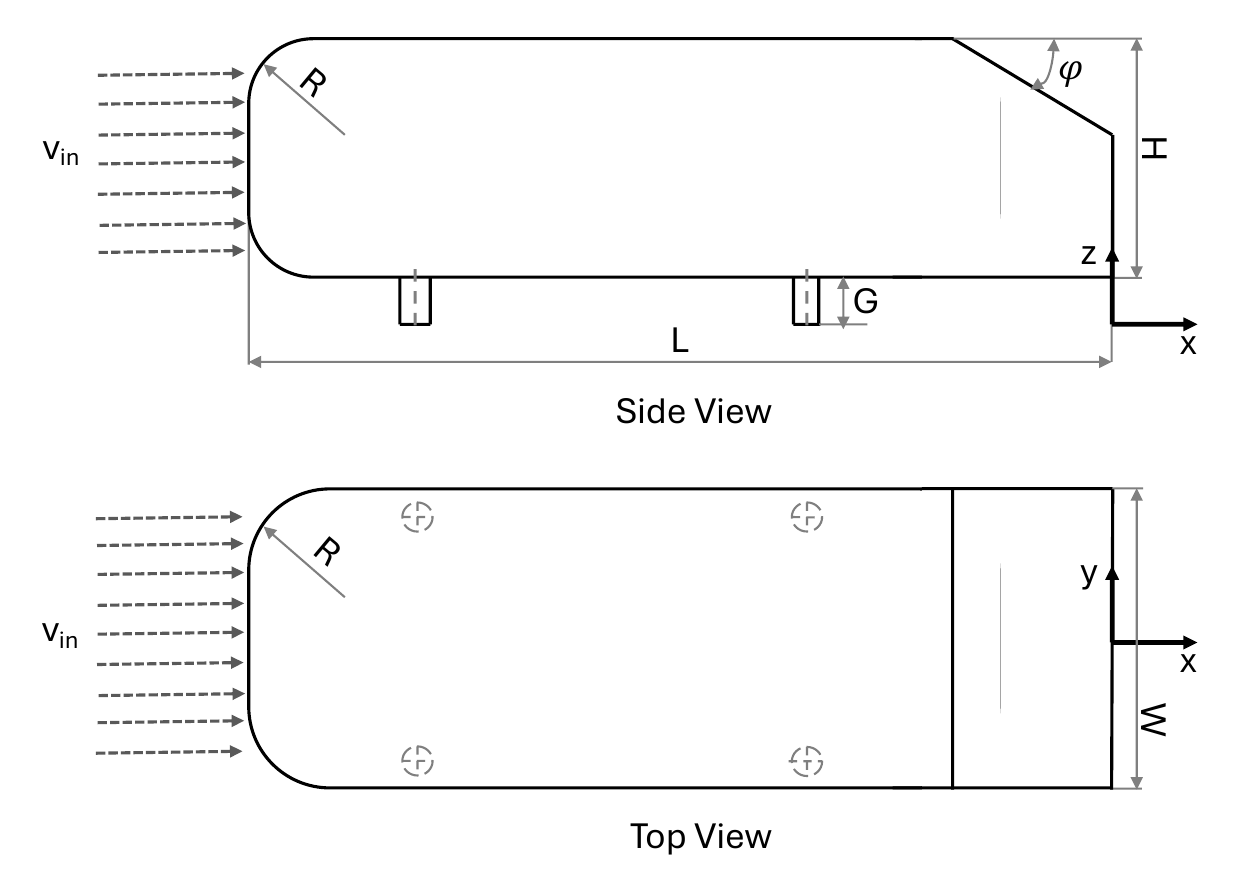}
         \caption{}
         \label{fig:ahmed_domain}
     \end{subfigure}
    \hfill
     \begin{subfigure}[b]{0.49\textwidth}
         \centering
         \includegraphics[width=0.8\textwidth]{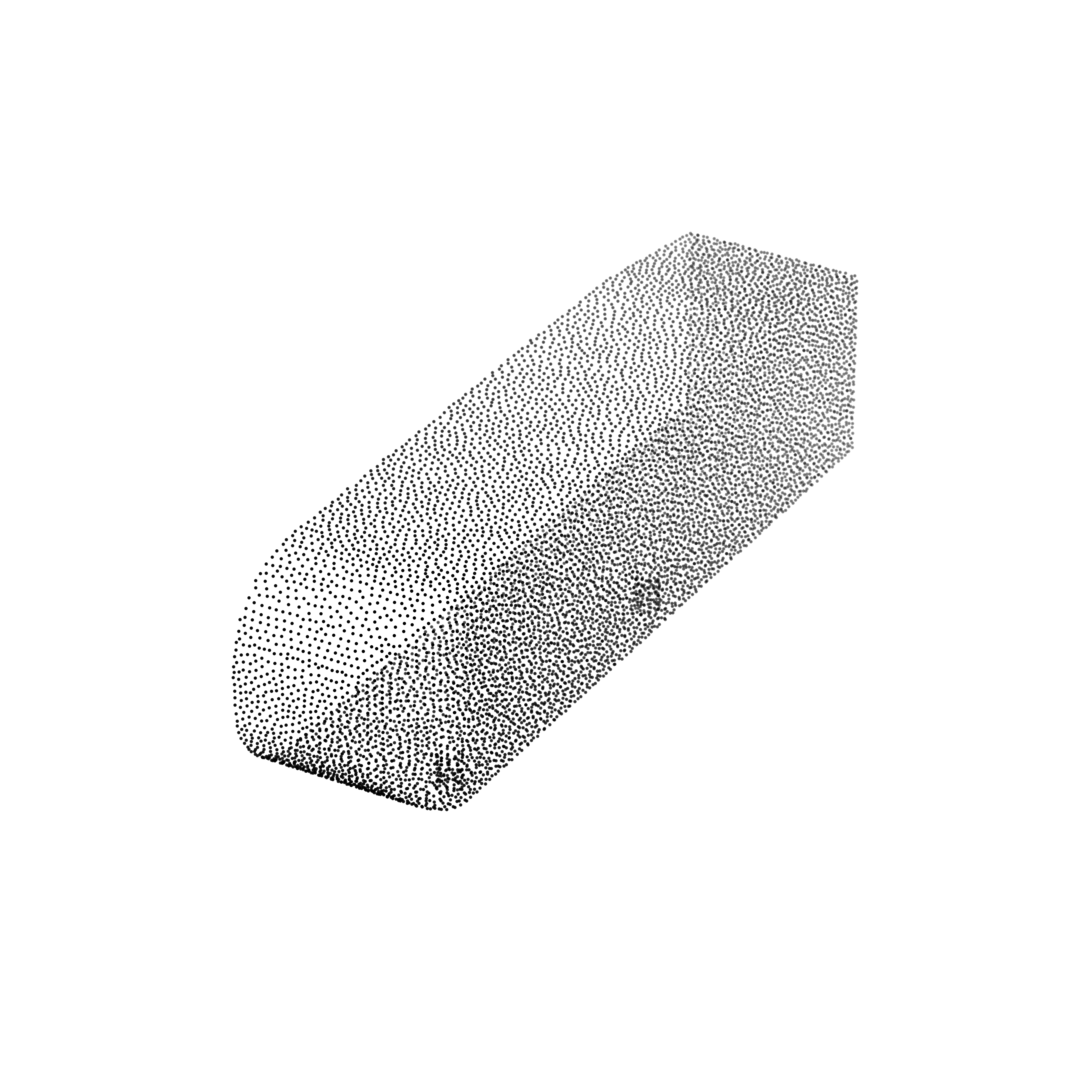}
         \caption{}
         \label{fig:ahmed_sample}
     \end{subfigure}
     \caption{Description of the Ahmed body used in the aerodynamics simulation, which includes (a) the geometry of Ahmed body of length, $L$, width, $W$, height, $H$, slant angle, $\varphi$, fillet radius, $R$ and ground clearance, $G$; and (b) a sample of low-resolution mesh (graph) from the training dataset for Ahmed body geometry,  containing 9,406 nodes.}
    \label{fig:ahmed_figs}
\end{figure}

In the final experiment, we consider the aerodynamics simulation on  3D Ahmed body geometries \cite{bayraktar2001experimental}, which are defined by six parameters: height ($H$), length ($L$), width ($W$), fillet radius ($R$), slant angle ($\varphi$), and ground clearance ($G$) as shown in Fig. \ref{fig:ahmed_domain}. The range of values of these parameters are given in Table \ref{table:param_values}. The simulations are performed with inlet velocities ($v_\text{in}$) ranging from $10 m/s$ to $70 m/s$, which correspond to Reynolds numbers between $4.35 \times 10^5$ and $6.82 \times 10^6$. This dataset is provided by the NVIDIA Modulus \cite{hennigh2021nvidia} team. In this experiment, the focus is on simulating the car's surface pressure and the wall shear stress, which are essential to calculate the drag coefficient, which reflects the vehicle’s aerodynamic efficiency of a vehicle. A lower value of the drag coefficient indicates a smoother passage through air or any fluid medium. This is an industry level problem with the low-fidelity meshes having an average of 8,600 nodes, medium-fidelity meshes with around 38,000 nodes and the high-fidelity meshes having an average of 69,000 nodes. Fig. \ref{fig:ahmed_sample} shows a sample from the training data for an Ahmed geometry with low mesh resolution.


\begin{table}[!ht]
\begin{center}
\caption{The range of values of geometry parameters considered for Ahmed body dataset.}
\begin{tabular}{lc}
\toprule
Parameter & Range  \\
\midrule
Length ($L$) & $700-1400$ m    \\
Width ($W$) & $250-550$ m    \\
Height ($H$) & $200-400$ m   \\
Fillet Radius ($R$) & $30-90$ m    \\
slant Angle ($\varphi$) & $0-40$   \\
Ground Clearance ($G$) & $80-120$ m\\
\bottomrule
\end{tabular}
\label{table:param_values}
\end{center}
\end{table}



We compare the performance of proposed GNN architectures with the benchmark models, used in the previous two examples, in predicting the distribution of pressure ($\bm \rho$) and wall shear stress components ($\bm \tau_x$, $\bm \tau_y$, $\bm \tau_z$) using relative L2-error defined as

\begin{equation}
    \begin{aligned}
    e_{\ell_2} = \dfrac{ \|\widehat{\bm f} - \bm f \|_2}{\|\bm f\|_2},
\label{relative_error_l2_ahmned}
    \end{aligned}
\end{equation}
where $\bm f$ is the output vector, $\bm f = [\bm \rho, \bm \tau_x, \bm \tau_y, \bm \tau_z]$, and $\widehat{\bm f}$ is the GNN prediction of $\bm f$. In this experiment, the training dataset for Transfer Learning-3, MF-UNet Lite-3 and MF-UNet-3 consists of 153 low-, medium- and high-resolution graphs, that for Transfer Learning-2, MF-UNet Lite-2 and MF-UNet-2 consists of 181 low- and high-resolution graphs and single-fidelity GNN consists of 240 high-resolution graphs. This is done to equalize the data generation time for all the models.  Additionally, the same architecture and hyper-parameters from the previous experiment are utilized for all the models in this experiment. However, in this experiment, after a number of trials,  we chose 12 GN blocks, with the coupling of node attributes for the proposed architectures done after the sixth GN block. Also, in this experiment, we consider $k=5$ in the $k$-nearest neighbors algorithm for mapping nodes between graphs of different resolutions. 



\begin{figure*}
     \centering
     \begin{subfigure}[t]{0.3\textwidth}
\vbox{
\vspace*{0.2em}%
\centering{
         \includegraphics[width=\textwidth]{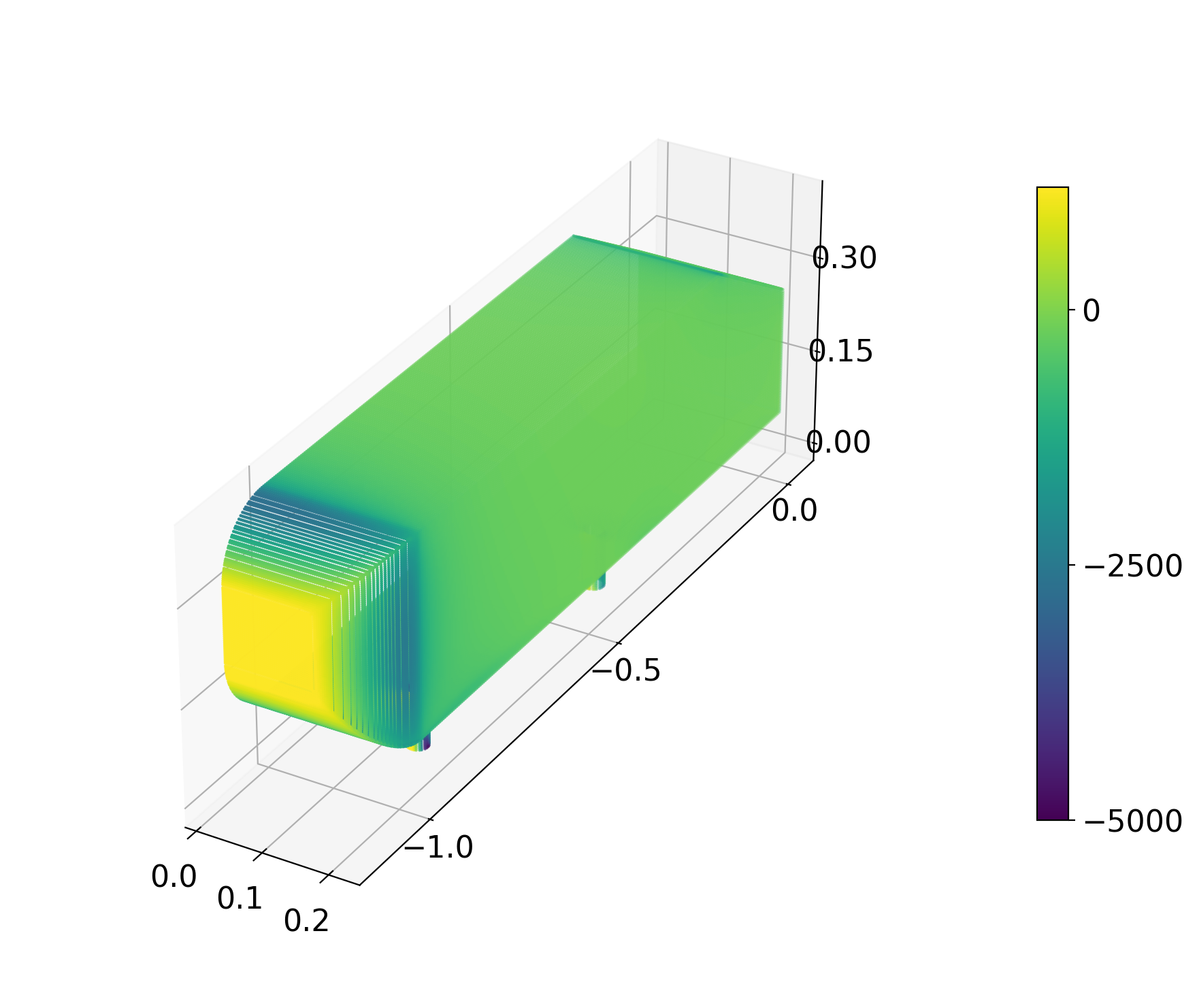}
         \caption{$\bm \rho$}
         \label{fig:exact_pressure}
         }%
\vspace*{0.2em}
}%
     \end{subfigure}
     \hfill
     \begin{subfigure}[t]{0.3\textwidth}
\vbox{
\vspace*{0.2em}%
\centering{
         \includegraphics[width=\textwidth]{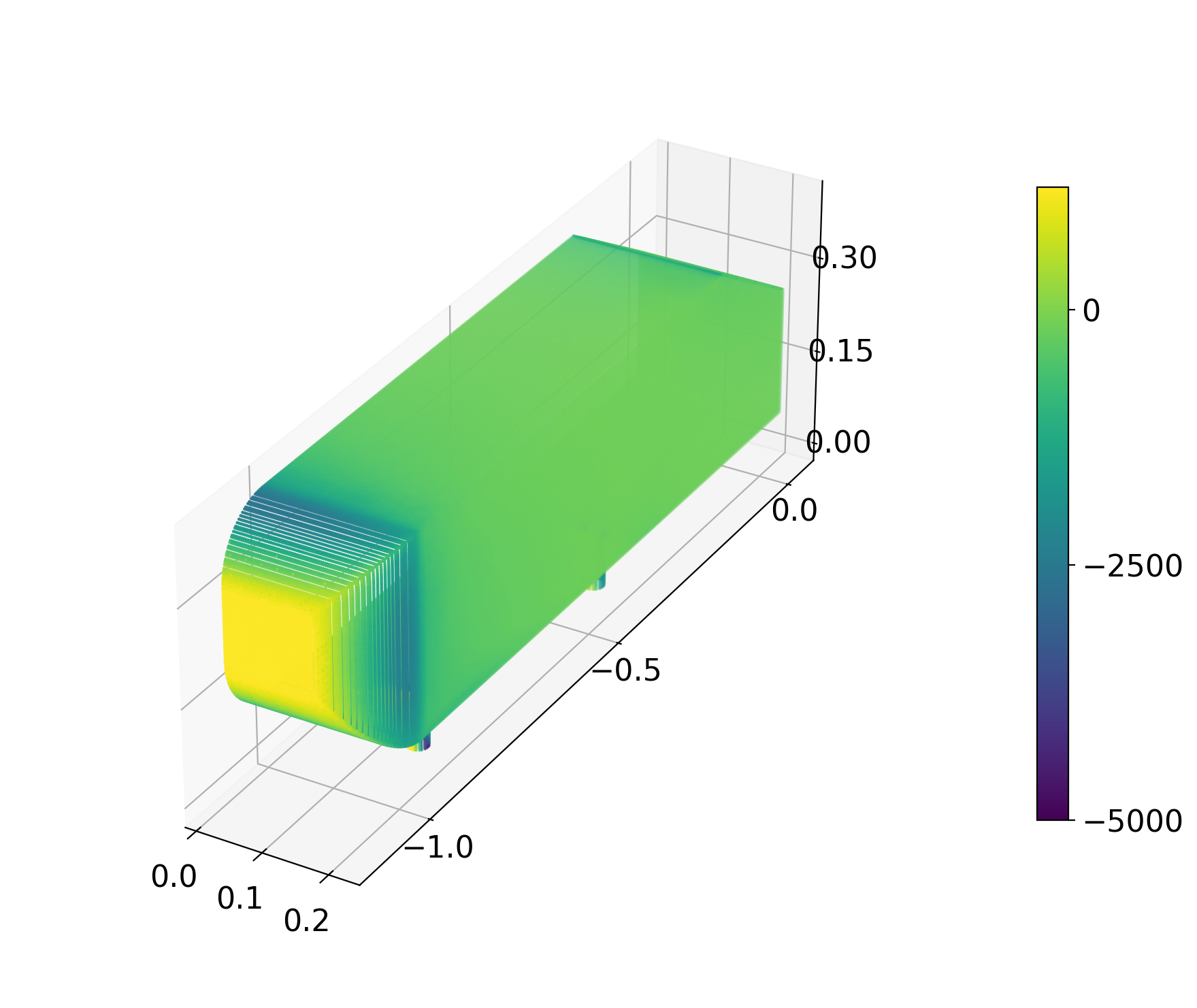}
         \caption{$\hat{\bm  \rho}$}
         \label{fig:pred_pressure}
         }%
\vspace*{0.2em}
}%
     \end{subfigure}
     \hfill
     \begin{subfigure}[t]{0.3\textwidth}
\vbox{
\vspace*{0.2em}%
\centering{
         \includegraphics[width=\textwidth]{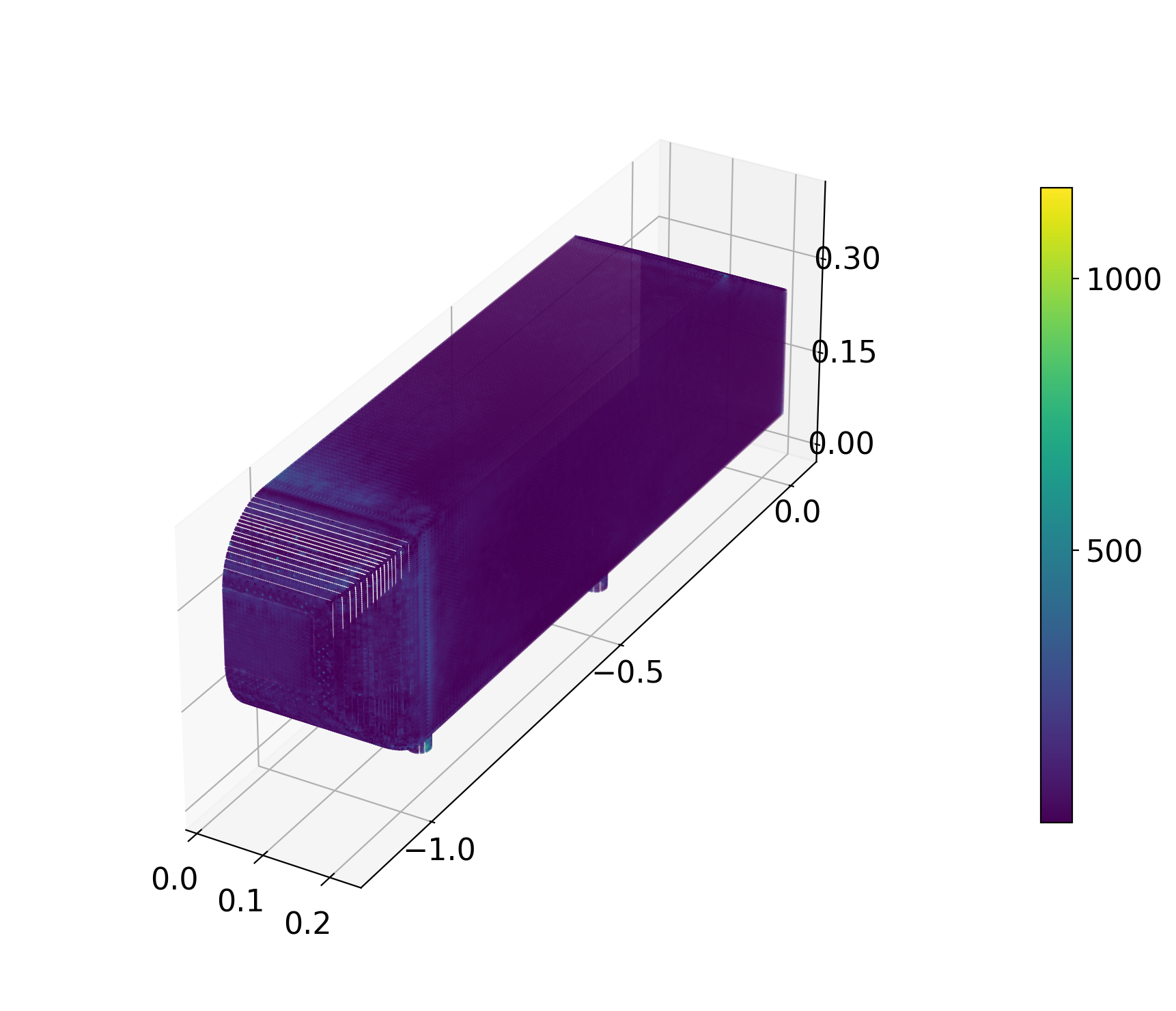}
         \caption{$|\bm \rho-\hat{\bm \rho}|$}
         \label{fig:diff_pressure}
         }%
\vspace*{0.2em}
}%
     \end{subfigure}
         \bigskip
     \begin{subfigure}[t]{0.3\textwidth}
\vbox{
\vspace*{0.2em}%
\centering{
         \includegraphics[width=\textwidth]{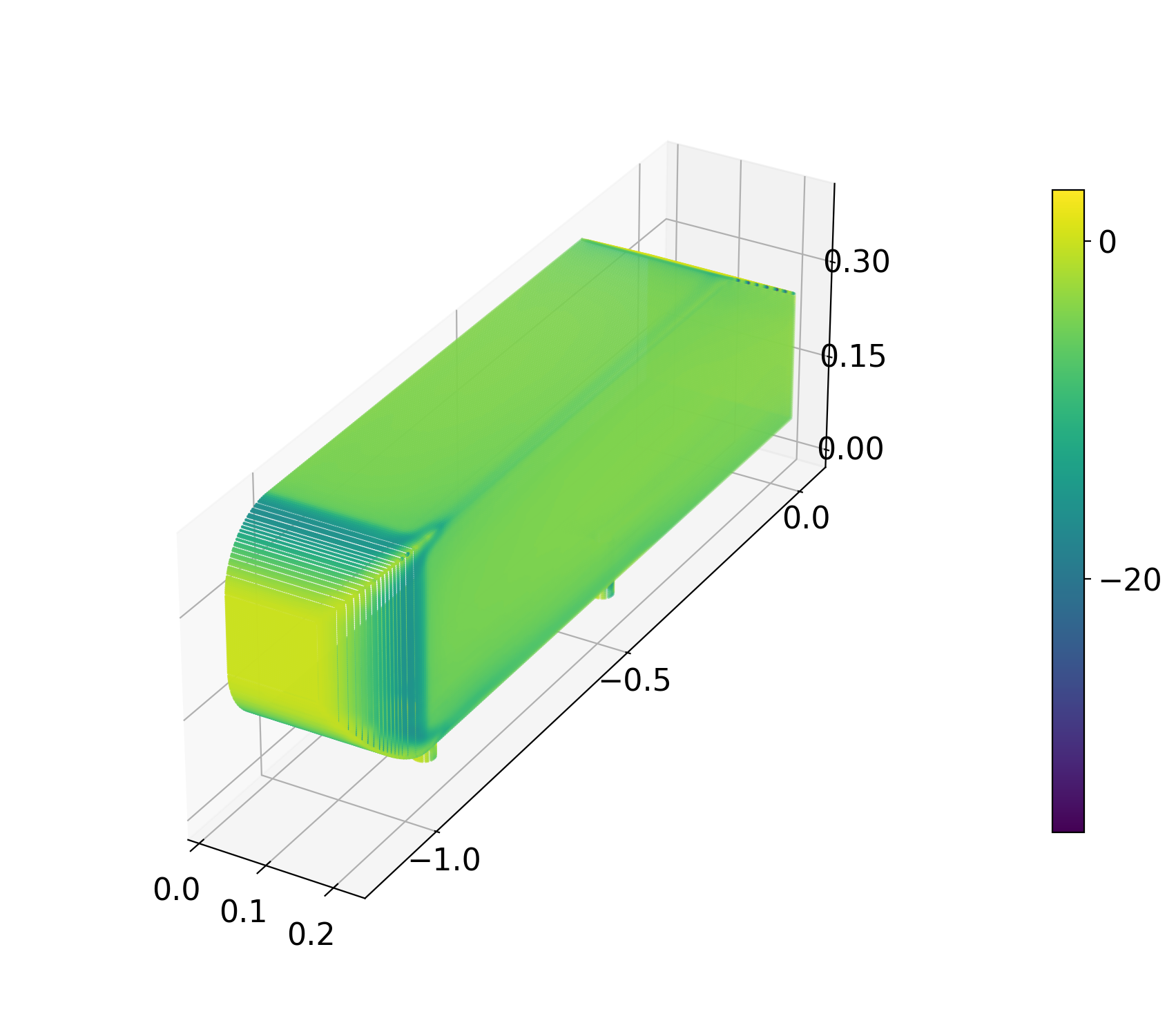}
         \caption{$\bm \tau_x$}
         \label{fig:exact_stress1}
         }%
\vspace*{0.2em}
}%
     \end{subfigure}
       \hfill
     \begin{subfigure}[t]{0.3\textwidth}
\vbox{
\vspace*{0.2em}%
\centering{
         \includegraphics[width=\textwidth]{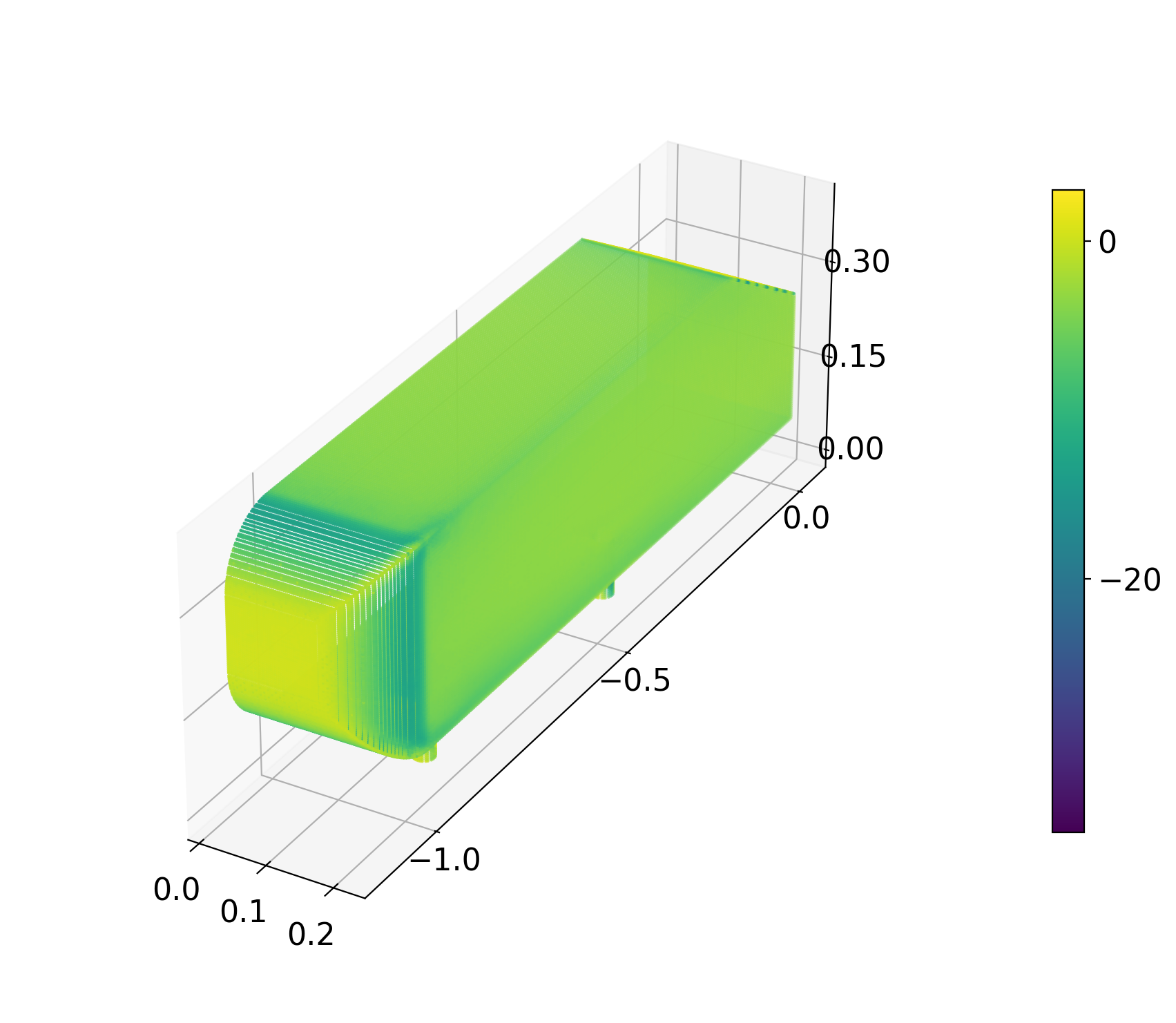}
         \caption{$\hat{\bm \tau}_x$}
         \label{fig:pred_stress1}
         }%
\vspace*{0.2em}
}%
     \end{subfigure}
     \hfill
\begin{subfigure}[t]{0.3\textwidth}
\vbox{
\vspace*{0.2em}%
\centering{
         \includegraphics[width=\textwidth]{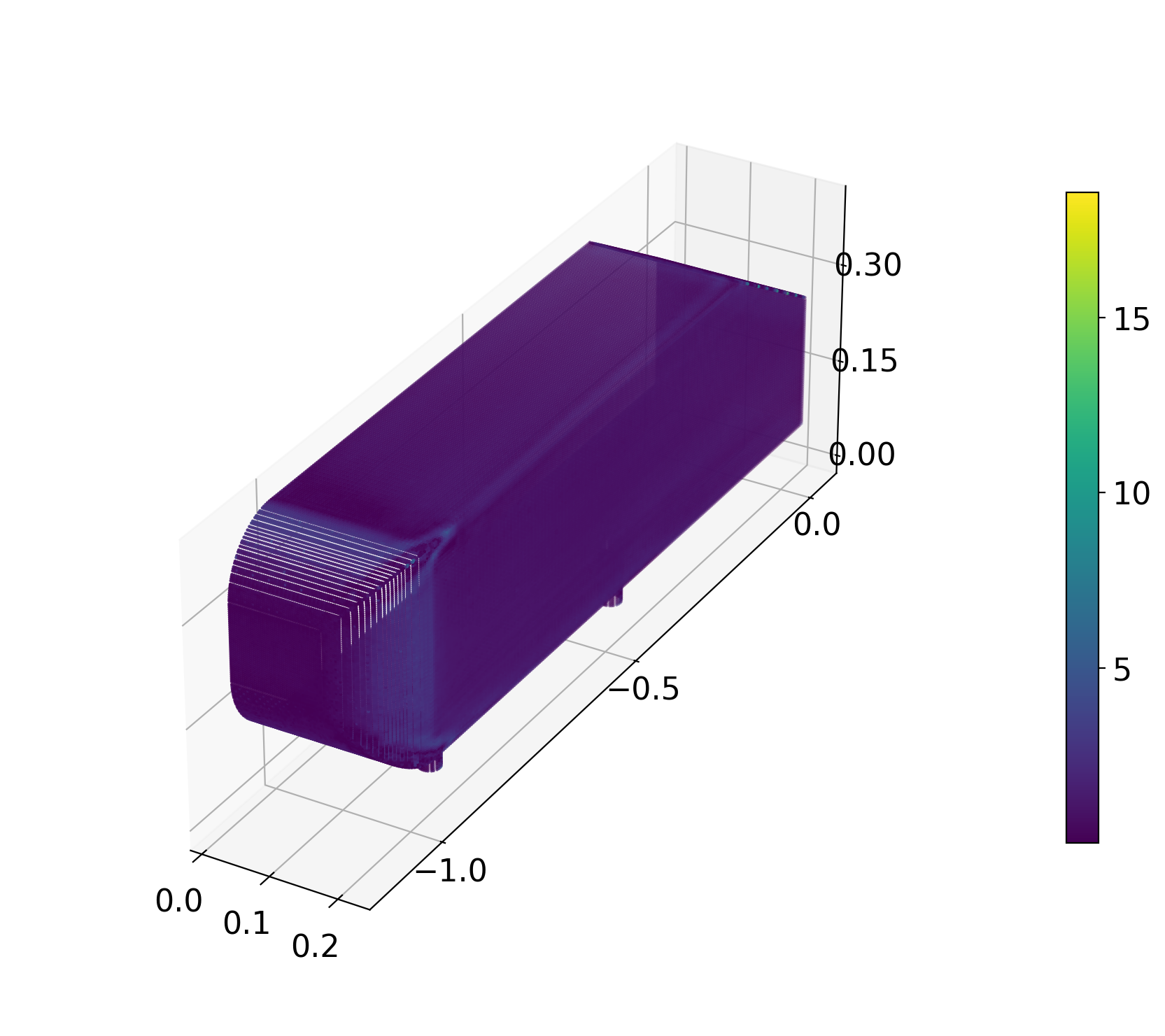}
         \caption{$|\bm \tau_x-\hat{\bm \tau}_x|$}
         \label{fig:diff_stress1}
         }%
\vspace*{0.2em}
}
     \end{subfigure}
         \bigskip
\begin{subfigure}[t]{0.3\textwidth}
\vbox{
\vspace*{0.2em}%
\centering{
         \includegraphics[width=\textwidth]{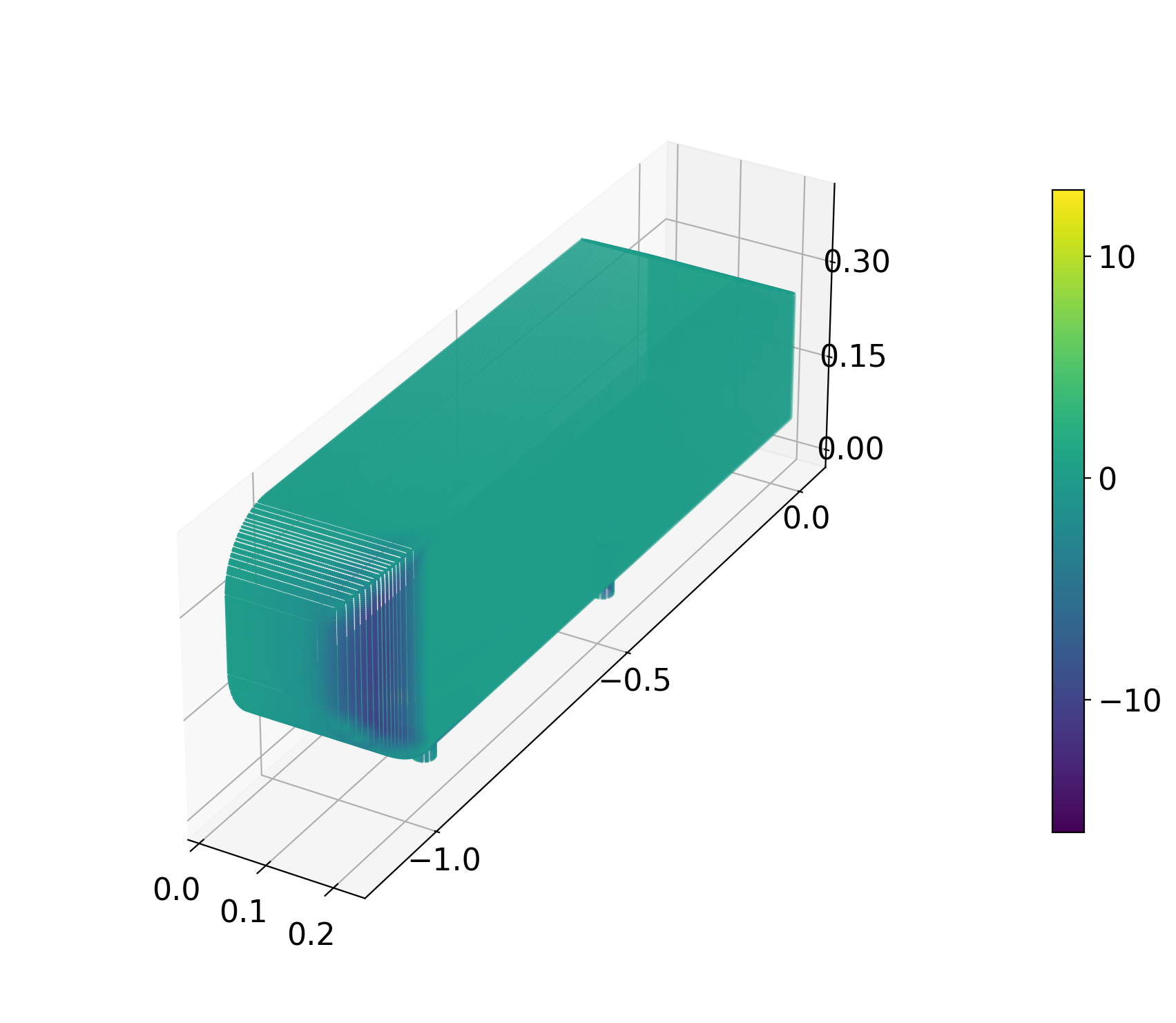}
         \caption{$\bm \tau_y$}
         \label{fig:exact_stress2}
         }%
\vspace*{0.2em}
}
     \end{subfigure}
          \hfill
\begin{subfigure}[t]{0.3\textwidth}
\vbox{
\vspace*{0.2em}%
\centering{
         \includegraphics[width=\textwidth]{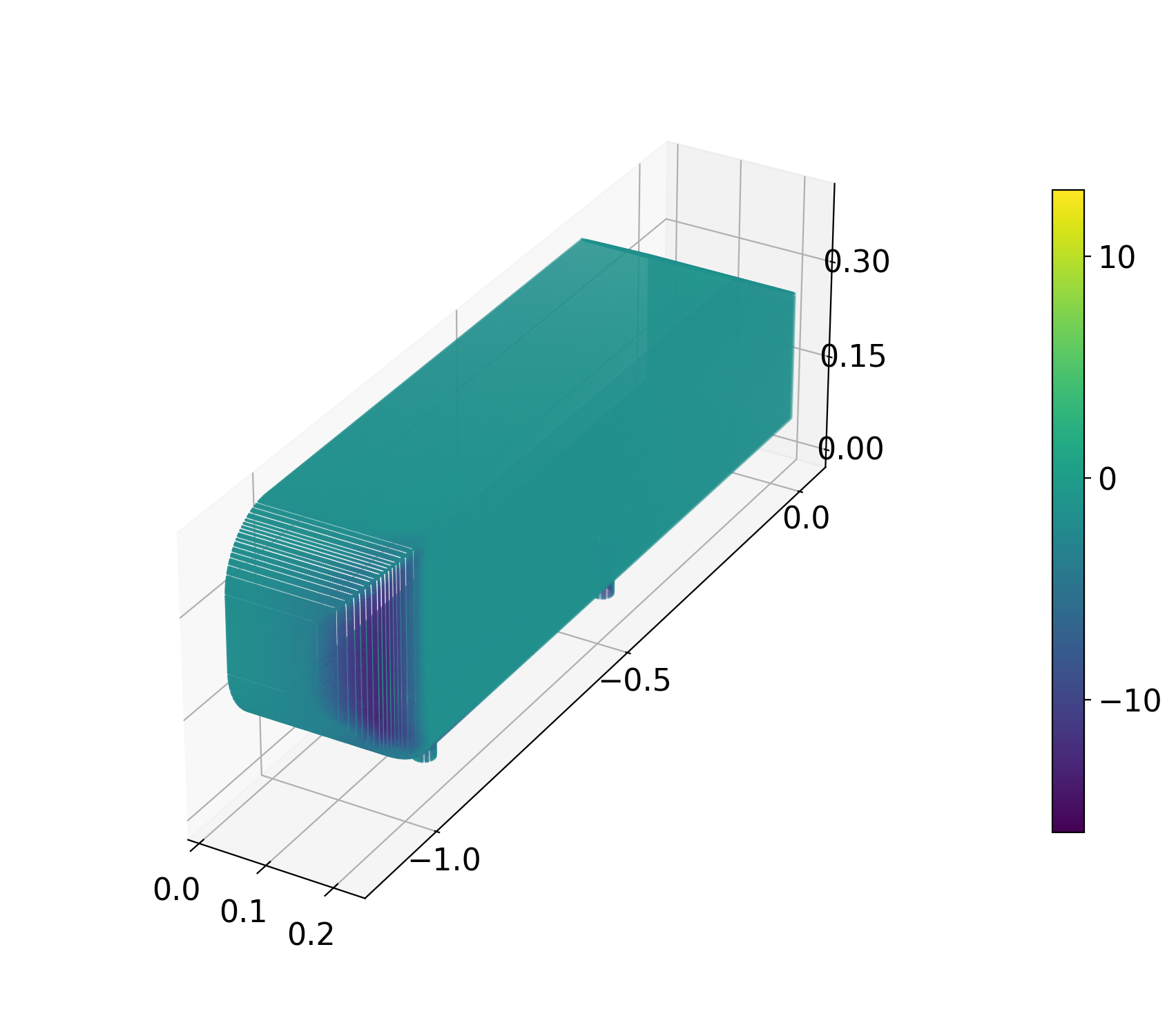}
         \caption{$\hat{\bm \tau}_y$}
         \label{fig:pred_stress2}
         }%
\vspace*{0.2em}
}
     \end{subfigure}
          \hfill
\begin{subfigure}[t]{0.3\textwidth}
\vbox{
\vspace*{0.2em}%
\centering{
         \includegraphics[width=\textwidth]{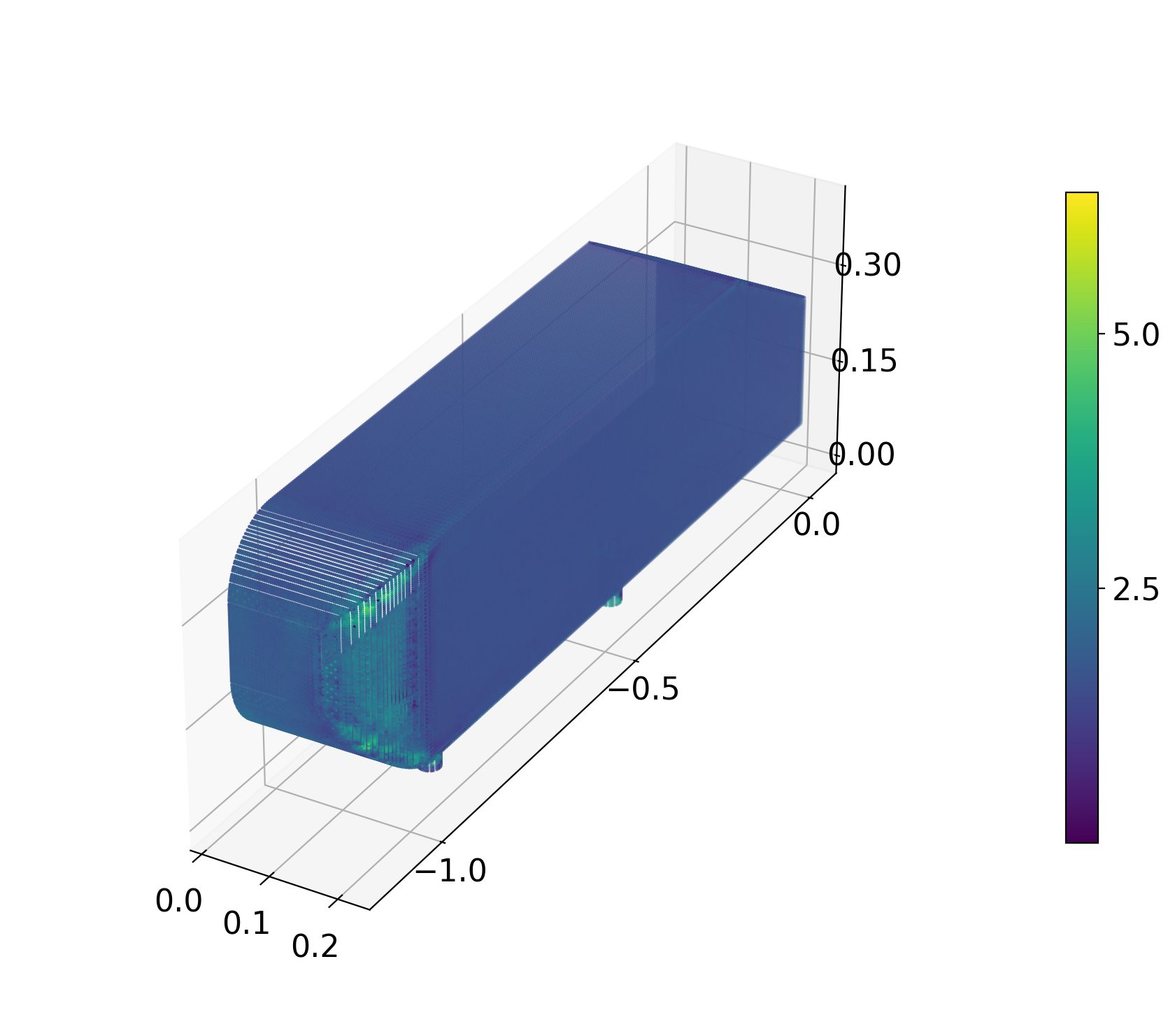}
         \caption{$|\bm \tau_y-\hat{\bm \tau}_y|$}
         \label{fig:diff_stress2}
         }%
\vspace*{0.2em}
}
     \end{subfigure}

         \bigskip
\begin{subfigure}[t]{0.3\textwidth}
\vbox{
\vspace*{0.2em}%
\centering{
         \includegraphics[width=\textwidth]{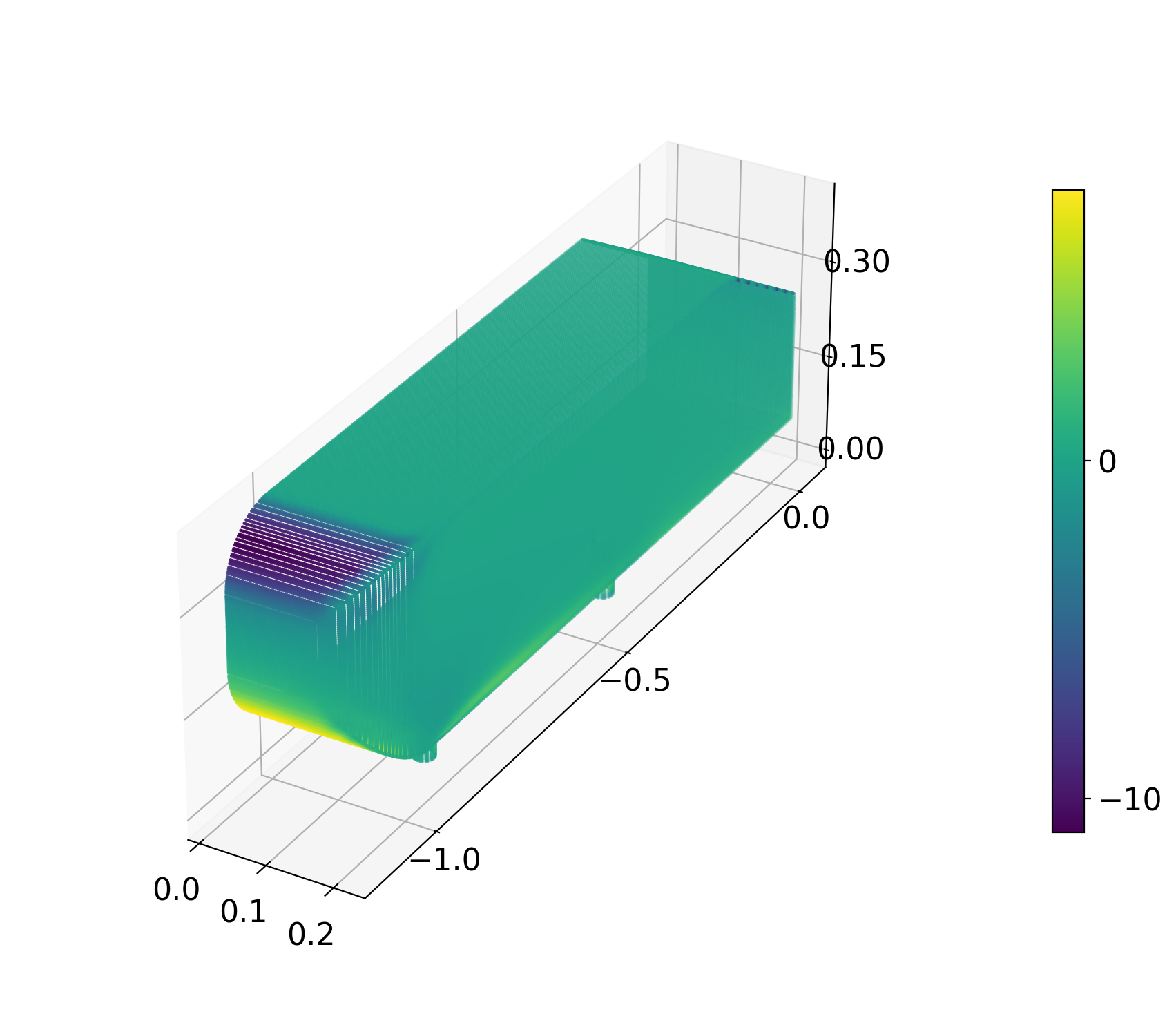}
         \caption{$\bm \tau_z$}
         \label{fig:exact_stress3}
         }%
\vspace*{0.2em}
}
     \end{subfigure}
          \hfill
\begin{subfigure}[t]{0.3\textwidth}
\vbox{
\vspace*{0.2em}%
\centering{
         \includegraphics[width=\textwidth]{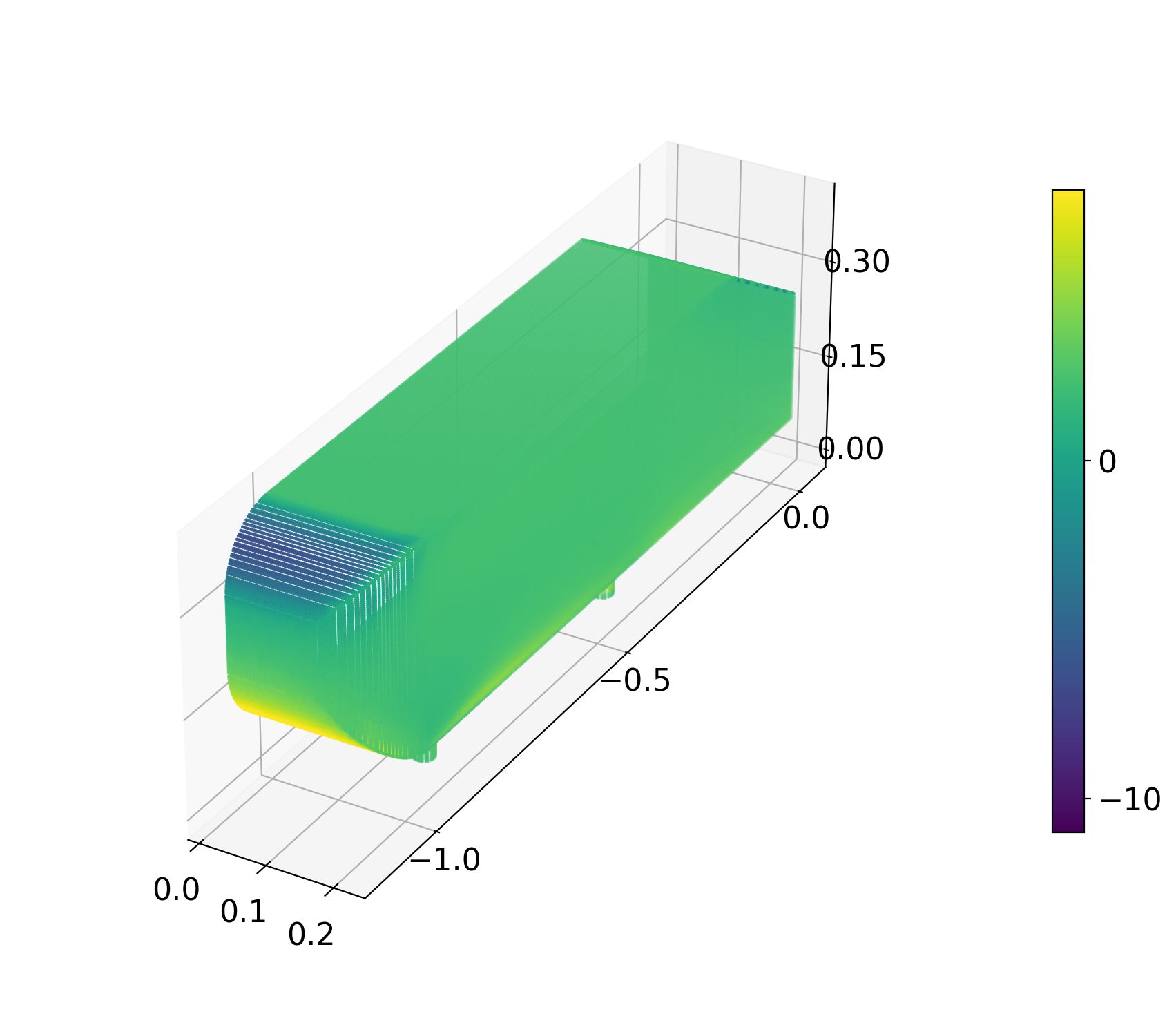}
         \caption{$\hat{\bm \tau}_z$}
         \label{fig:pred_stress3}
         }%
\vspace*{0.2em}
}
     \end{subfigure}
          \hfill
\begin{subfigure}[t]{0.3\textwidth}
\vbox{
\vspace*{0.2em}%
\centering{
         \includegraphics[width=\textwidth]{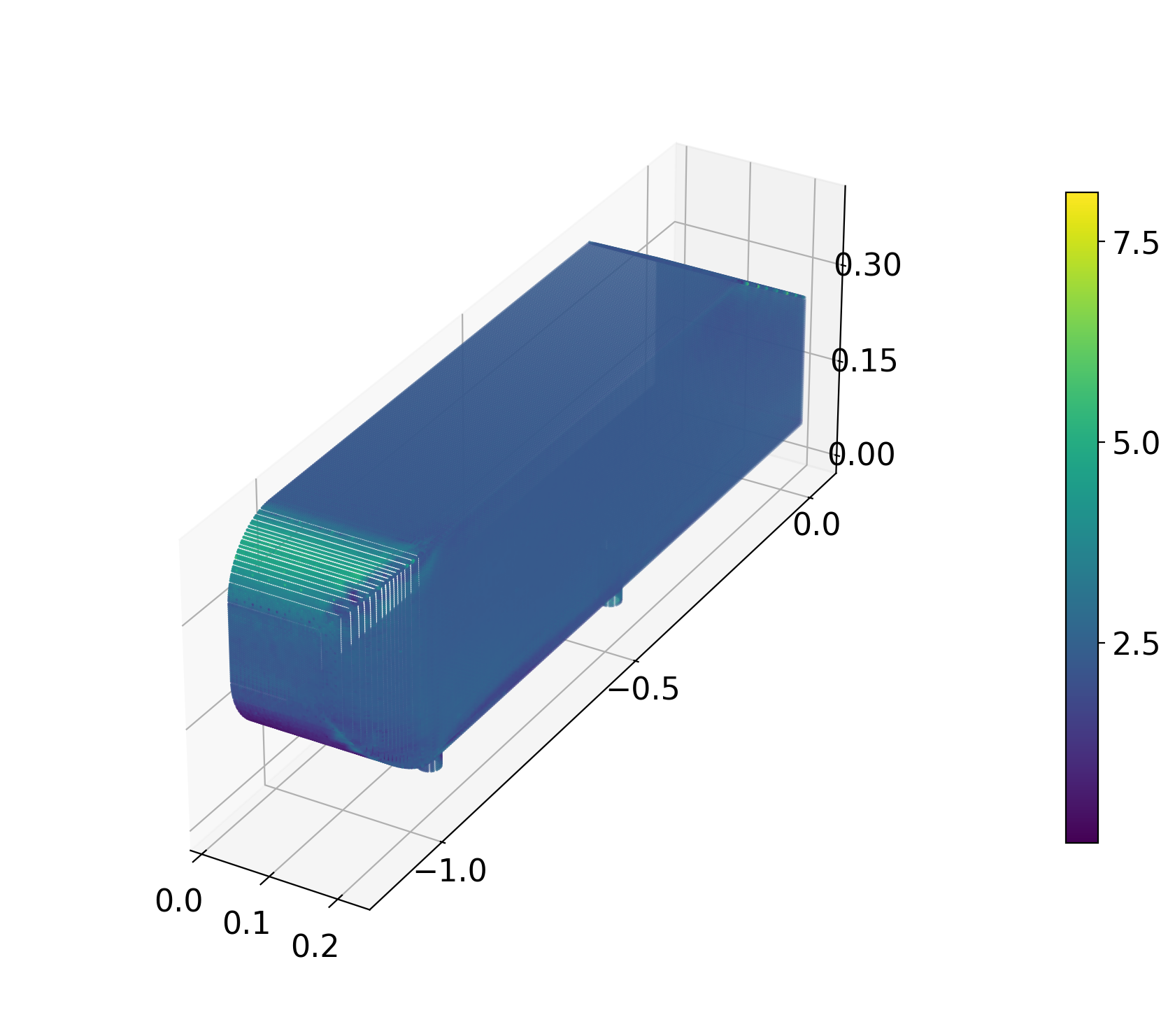}
         \caption{$|\bm \tau_z-\hat{\bm \tau}_z|$}
         \label{fig:diff_stress3}
         }%
\vspace*{0.2em}
}
     \end{subfigure}
    \caption{The exact and  predicted distributions of the pressure, $\bm \rho$, and wall shear stress in $x$, $y$, $z$ directions ($\bm \tau_x$, $\bm \tau_y$, $\bm \tau_z$), along with the absolute prediction error, for a representative high-resolution sample. The predictions are shown for the model, MF-UNet-3. }
    \label{fig:ahmed_sample_y}
\end{figure*}

\begin{table}[!ht]
\begin{center}
\caption{The relative L2-error as defined in Eq.~\eqref{relative_error_l2_ahmned} across the testing dataset for different models for evaluating the distribution of pressure and wall shear stress for the Ahmed dataset for high-resolution graphs. }
\begin{tabular}{lcr}
\toprule
Model & \# Parameters & $e_{l2}$\\
\midrule
Single Fidelity & 2,034,820 & 28.1\%  \\
Transfer Learning-2 & 2,034,820 &  20.0\% \\
Transfer Learning-3 & 2,034,820 &  19.6\%  \\
MF-UNet-2 & 2,034,820 & 15.0\% \\
MF-UNet-3 & 2,034,820 & \textbf{14.2\%} \\
MF-UNet Lite-2 & 2,034,820 & 16.2\%  \\
MF-UNet Lite-3 & 2,034,820 & 15.3\%  \\
\bottomrule
\end{tabular}
\label{table:ahmed_errors}
\end{center}
\end{table}

Table~\ref{table:ahmed_errors} presents a comparison of different GNN methods in terms of accuracy (relative L2-error) and number of model parameters for the Ahmed body test data set. All the multi-fidelity models, including the proposed architectures and the transfer learning method, outperform the single-fidelity GNN model, even when additional high-fidelity data are incorporated into the single-fidelity model. MF-UNet is the best performing model in terms of accuracy, with the three levels of fidelity consistently performing better than the two levels of fidelity across all the multi-fidelity models. Fig. \ref{fig:ahmed_sample_y} shows the exact and predicted distribution of the outputs, pressure ($\rho$) and wall shear stress in the three directions ($\tau_x$, $\tau_y$ and $\tau_z$) and the corresponding absolute prediction errors for a representative high-resolution graph for this model. These findings are in line with the observations from the previous two examples. This underscores the substantial benefits of multi-fidelity approaches in evaluating the responses for high-resolution graphs, particularly when enhanced with the information from lower-fidelity data during the training of the models.

\section{Discussion}
\label{discussion}
In this study, we introduced a novel approach, Multi-Fidelity U-Net (MF-UNet) and a its faster version, MF-UNet Lite, to integrate multi-fidelity modeling with GNNs for mesh-based simulation of PDEs. The proposed GNN architectures utilize  the strengths of different fidelities of data, by enabling flow of information between the data during the training, resulting in significant increase in the prediction performance for the high-fidelity meshes, while reducing the computational requirements of high-fidelity data generation and model training. The efficacy of the proposed models is demonstrated across a variety of experiments, ranging from 2D displacement and stress predictions to complex 3D CFD simulations. The results demonstrate that the proposed multi-fidelity GNN models achieve higher accuracy compared to single-fidelity and transfer learning models while maintaining the same model complexity. The proposed multi-fidelity frameworks for GNNs not only set a new benchmark for computational efficiency in solving PDEs for high-fidelity meshes but also opens avenues for its application in real-world problems where high-resolution simulations are generally computationally prohibitive. The demonstrated advantage of our proposed GNN architectures across  various examples presented in this study, highlight the applicability of this multi-fidelity framework for a wide array of problems. The inherent ability of GNNs to manage unstructured and complex meshes and domains and the use of low-fidelity data to enrich the capability of GNNs to extract more information of the physical system to enhance the predictability of the model for high-fidelity data make the proposed methods suitable and generalizable for complex, industry-level problems with varying geometry and boundary conditions.  Further research is needed to extend  the proposed models to time-dependent PDEs, and to investigate better upsampling techniques for passing information across different levels of fidelity. 

\section*{Acknowledgement}
The authors would like to thank Mehdi Taghizadeh and Negin Alemazkoor in providing the code to generate the data for the 2D plate with notches and variable hole and Mohammad Amin Nabian and NVIDIA for providing the Ahmed body geometry data. The authors also acknowledge the support from Nolan Black in providing guidance on generating the dataset for the 2D cantilever beam example. 

\bibliographystyle{plainnat}
\bibliography{references}

\begin{thebibliography}{55}
\providecommand{\natexlab}[1]{#1}
\providecommand{\url}[1]{\texttt{#1}}
\expandafter\ifx\csname urlstyle\endcsname\relax
  \providecommand{\doi}[1]{doi: #1}\else
  \providecommand{\doi}{doi: \begingroup \urlstyle{rm}\Url}\fi

\bibitem[Aln{\ae}s et~al.(2015)Aln{\ae}s, Blechta, Hake, Johansson, Kehlet, Logg, Richardson, Ring, Rognes, and Wells]{alnaes2015fenics}
Martin Aln{\ae}s, Jan Blechta, Johan Hake, August Johansson, Benjamin Kehlet, Anders Logg, Chris Richardson, Johannes Ring, Marie~E Rognes, and Garth~N Wells.
\newblock The fenics project version 1.5.
\newblock \emph{Archive of numerical software}, 3\penalty0 (100), 2015.

\bibitem[Battaglia et~al.(2018)Battaglia, Hamrick, Bapst, Sanchez-Gonzalez, Zambaldi, Malinowski, Tacchetti, Raposo, Santoro, Faulkner, et~al.]{battaglia2018relational}
Peter~W Battaglia, Jessica~B Hamrick, Victor Bapst, Alvaro Sanchez-Gonzalez, Vinicius Zambaldi, Mateusz Malinowski, Andrea Tacchetti, David Raposo, Adam Santoro, Ryan Faulkner, et~al.
\newblock Relational inductive biases, deep learning, and graph networks.
\newblock \emph{arXiv preprint arXiv:1806.01261}, 2018.

\bibitem[Bayraktar et~al.(2001)Bayraktar, Landman, and Baysal]{bayraktar2001experimental}
Ilhan Bayraktar, Drew Landman, and Oktay Baysal.
\newblock Experimental and computational investigation of ahmed body for ground vehicle aerodynamics.
\newblock \emph{SAE transactions}, pages 321--331, 2001.

\bibitem[Black and Najafi(2022)]{black2022learning}
Nolan Black and Ahmad~R Najafi.
\newblock Learning finite element convergence with the multi-fidelity graph neural network.
\newblock \emph{Computer Methods in Applied Mechanics and Engineering}, 397:\penalty0 115120, 2022.

\bibitem[Chakraborty(2021)]{chakraborty2021transfer}
Souvik Chakraborty.
\newblock Transfer learning based multi-fidelity physics informed deep neural network.
\newblock \emph{Journal of Computational Physics}, 426:\penalty0 109942, 2021.

\bibitem[Chen et~al.(2021)Chen, Zuo, Ye, Li, and Ong]{chen2021learning}
Chi Chen, Yunxing Zuo, Weike Ye, Xiangguo Li, and Shyue~Ping Ong.
\newblock Learning properties of ordered and disordered materials from multi-fidelity data.
\newblock \emph{Nature Computational Science}, 1\penalty0 (1):\penalty0 46--53, 2021.

\bibitem[Deshpande et~al.(2024)Deshpande, Bordas, and Lengiewicz]{deshpande2024magnet}
Saurabh Deshpande, St{\'e}phane~PA Bordas, and Jakub Lengiewicz.
\newblock Magnet: A graph u-net architecture for mesh-based simulations.
\newblock \emph{Engineering Applications of Artificial Intelligence}, 133:\penalty0 108055, 2024.

\bibitem[Donon et~al.(2019)Donon, Donnot, Guyon, and Marot]{donon2019graph}
Balthazar Donon, Benjamin Donnot, Isabelle Guyon, and Antoine Marot.
\newblock Graph neural solver for power systems.
\newblock In \emph{2019 international joint conference on neural networks (ijcnn)}, pages 1--8. IEEE, 2019.

\bibitem[Fern{\'a}ndez-Godino(2016)]{fernandez2016review}
M~Giselle Fern{\'a}ndez-Godino.
\newblock Review of multi-fidelity models.
\newblock \emph{arXiv preprint arXiv:1609.07196}, 2016.

\bibitem[Fortunato et~al.(2022)Fortunato, Pfaff, Wirnsberger, Pritzel, and Battaglia]{multiscale}
Meire Fortunato, Tobias Pfaff, Peter Wirnsberger, Alexander Pritzel, and Peter Battaglia.
\newblock Multiscale meshgraphnets.
\newblock \emph{arXiv preprint arXiv:2210.00612}, 2022.

\bibitem[Giselle Fern{\'a}ndez-Godino et~al.(2019)Giselle Fern{\'a}ndez-Godino, Park, Kim, and Haftka]{giselle2019issues}
M~Giselle Fern{\'a}ndez-Godino, Chanyoung Park, Nam~H Kim, and Raphael~T Haftka.
\newblock Issues in deciding whether to use multifidelity surrogates.
\newblock \emph{Aiaa Journal}, 57\penalty0 (5):\penalty0 2039--2054, 2019.

\bibitem[Gladstone et~al.(2022)Gladstone, Nabian, and Meidani]{gladstone2022fo}
Rini~J Gladstone, Mohammad~A Nabian, and Hadi Meidani.
\newblock Fo-pinns: A first-order formulation for physics informed neural networks.
\newblock \emph{arXiv preprint arXiv:2210.14320}, 2022.

\bibitem[Gladstone et~al.(2021)Gladstone, Nabian, Keshavarzzadeh, and Meidani]{gladstone2021robust}
Rini~Jasmine Gladstone, Mohammad~Amin Nabian, Vahid Keshavarzzadeh, and Hadi Meidani.
\newblock Robust topology optimization using variational autoencoders.
\newblock \emph{arXiv preprint arXiv:2107.10661}, 2021.

\bibitem[Gladstone et~al.(2023)Gladstone, Rahmani, Suryakumar, Meidani, D'Elia, and Zareei]{gladstone2023gnnbased}
Rini~Jasmine Gladstone, Helia Rahmani, Vishvas Suryakumar, Hadi Meidani, Marta D'Elia, and Ahmad Zareei.
\newblock Gnn-based physics solver for time-independent pdes, 2023.

\bibitem[Gladstone et~al.(2024)Gladstone, Rahmani, Suryakumar, Meidani, D’Elia, and Zareei]{gladstone2024mesh}
Rini~Jasmine Gladstone, Helia Rahmani, Vishvas Suryakumar, Hadi Meidani, Marta D’Elia, and Ahmad Zareei.
\newblock Mesh-based gnn surrogates for time-independent pdes.
\newblock \emph{Scientific Reports}, 14\penalty0 (1):\penalty0 3394, 2024.

\bibitem[Gorodetsky et~al.(2021)Gorodetsky, Jakeman, and Geraci]{gorodetsky2021mfnets}
Alex~A Gorodetsky, John~D Jakeman, and Gianluca Geraci.
\newblock Mfnets: data efficient all-at-once learning of multifidelity surrogates as directed networks of information sources.
\newblock \emph{Computational Mechanics}, 68\penalty0 (4):\penalty0 741--758, 2021.

\bibitem[Hamilton et~al.(2017)Hamilton, Ying, and Leskovec]{hamilton2017inductive}
Will Hamilton, Zhitao Ying, and Jure Leskovec.
\newblock Inductive representation learning on large graphs.
\newblock \emph{Advances in neural information processing systems}, 30, 2017.

\bibitem[He et~al.(2020)He, Qian, Zhao, and Wang]{he2020multi}
Lei He, Weiqi Qian, Tun Zhao, and Qing Wang.
\newblock Multi-fidelity aerodynamic data fusion with a deep neural network modeling method.
\newblock \emph{Entropy}, 22\penalty0 (9):\penalty0 1022, 2020.

\bibitem[Hennigh et~al.(2021)Hennigh, Narasimhan, Nabian, Subramaniam, Tangsali, Fang, Rietmann, Byeon, and Choudhry]{hennigh2021nvidia}
Oliver Hennigh, Susheela Narasimhan, Mohammad~Amin Nabian, Akshay Subramaniam, Kaustubh Tangsali, Zhiwei Fang, Max Rietmann, Wonmin Byeon, and Sanjay Choudhry.
\newblock Nvidia simnet™: An ai-accelerated multi-physics simulation framework.
\newblock In \emph{International Conference on Computational Science}, pages 447--461. Springer, 2021.

\bibitem[Ju et~al.(2020)Ju, Farrell, Calafiura, Murnane, Gray, Klijnsma, Pedro, Cerati, Kowalkowski, Perdue, et~al.]{ju2020graph}
Xiangyang Ju, Steven Farrell, Paolo Calafiura, Daniel Murnane, Lindsey Gray, Thomas Klijnsma, Kevin Pedro, Giuseppe Cerati, Jim Kowalkowski, Gabriel Perdue, et~al.
\newblock Graph neural networks for particle reconstruction in high energy physics detectors.
\newblock \emph{arXiv preprint arXiv:2003.11603}, 2020.

\bibitem[Kaszynski(2021)]{alexander_kaszynski_2020_4009467}
Alexander Kaszynski.
\newblock {pyansys: Pythonic interface to MAPDL}, November 2021.
\newblock URL \url{https://doi.org/10.5281/zenodo.4009466}.

\bibitem[Khoo et~al.(2021)Khoo, Lu, and Ying]{khoo2021solving}
Yuehaw Khoo, Jianfeng Lu, and Lexing Ying.
\newblock Solving parametric pde problems with artificial neural networks.
\newblock \emph{European Journal of Applied Mathematics}, 32\penalty0 (3):\penalty0 421--435, 2021.

\bibitem[Kipf and Welling(2016)]{kipf2016semi}
Thomas~N Kipf and Max Welling.
\newblock Semi-supervised classification with graph convolutional networks.
\newblock \emph{arXiv preprint arXiv:1609.02907}, 2016.

\bibitem[Lagaris et~al.(1998)Lagaris, Likas, and Fotiadis]{lagaris1998artificial}
Isaac~E Lagaris, Aristidis Likas, and Dimitrios~I Fotiadis.
\newblock Artificial neural networks for solving ordinary and partial differential equations.
\newblock \emph{IEEE transactions on neural networks}, 9\penalty0 (5):\penalty0 987--1000, 1998.

\bibitem[Li et~al.(2023)Li, Li, Liu, Zhang, and Xie]{li2023multi}
Jinxing Li, Yunzhu Li, Tianyuan Liu, Di~Zhang, and Yonghui Xie.
\newblock Multi-fidelity graph neural network for flow field data fusion of turbomachinery.
\newblock \emph{Energy}, 285:\penalty0 129405, 2023.

\bibitem[Li et~al.(2020{\natexlab{a}})Li, Kovachki, Azizzadenesheli, Liu, Bhattacharya, Stuart, and Anandkumar]{li2020fourier}
Zongyi Li, Nikola Kovachki, Kamyar Azizzadenesheli, Burigede Liu, Kaushik Bhattacharya, Andrew Stuart, and Anima Anandkumar.
\newblock Fourier neural operator for parametric partial differential equations.
\newblock \emph{arXiv preprint arXiv:2010.08895}, 2020{\natexlab{a}}.

\bibitem[Li et~al.(2020{\natexlab{b}})Li, Kovachki, Azizzadenesheli, Liu, Bhattacharya, Stuart, and Anandkumar]{li2020neural}
Zongyi Li, Nikola Kovachki, Kamyar Azizzadenesheli, Burigede Liu, Kaushik Bhattacharya, Andrew Stuart, and Anima Anandkumar.
\newblock Neural operator: Graph kernel network for partial differential equations.
\newblock \emph{arXiv preprint arXiv:2003.03485}, 2020{\natexlab{b}}.

\bibitem[Li et~al.(2020{\natexlab{c}})Li, Kovachki, Azizzadenesheli, Liu, Stuart, Bhattacharya, and Anandkumar]{li2020multipole}
Zongyi Li, Nikola Kovachki, Kamyar Azizzadenesheli, Burigede Liu, Andrew Stuart, Kaushik Bhattacharya, and Anima Anandkumar.
\newblock Multipole graph neural operator for parametric partial differential equations.
\newblock \emph{Advances in Neural Information Processing Systems}, 33:\penalty0 6755--6766, 2020{\natexlab{c}}.

\bibitem[Liu et~al.(2022)Liu, Yagoubi, Danan, and Schoenauer]{liu2022multi}
Wenzhuo Liu, Mouadh Yagoubi, David Danan, and Marc Schoenauer.
\newblock Multi-fidelity transfer learning for accurate data-based pde approximation.
\newblock In \emph{NeurIPS 2022-Workshop on Machine Learning and the Physical Sciences}, 2022.

\bibitem[Loshchilov et~al.(2017)Loshchilov, Hutter, et~al.]{loshchilov2017fixing}
Ilya Loshchilov, Frank Hutter, et~al.
\newblock Fixing weight decay regularization in adam.
\newblock \emph{arXiv preprint arXiv:1711.05101}, 5, 2017.

\bibitem[Lu et~al.(2021)Lu, Jin, Pang, Zhang, and Karniadakis]{lu2021learning}
Lu~Lu, Pengzhan Jin, Guofei Pang, Zhongqiang Zhang, and George~Em Karniadakis.
\newblock Learning nonlinear operators via deeponet based on the universal approximation theorem of operators.
\newblock \emph{Nature machine intelligence}, 3\penalty0 (3):\penalty0 218--229, 2021.

\bibitem[Mahmoudabadbozchelou et~al.(2021)Mahmoudabadbozchelou, Caggioni, Shahsavari, Hartt, Em~Karniadakis, and Jamali]{mahmoudabadbozchelou2021data}
Mohammadamin Mahmoudabadbozchelou, Marco Caggioni, Setareh Shahsavari, William~H Hartt, George Em~Karniadakis, and Safa Jamali.
\newblock Data-driven physics-informed constitutive metamodeling of complex fluids: A multifidelity neural network (mfnn) framework.
\newblock \emph{Journal of Rheology}, 65\penalty0 (2):\penalty0 179--198, 2021.

\bibitem[Meng and Karniadakis(2020)]{meng2020composite}
Xuhui Meng and George~Em Karniadakis.
\newblock A composite neural network that learns from multi-fidelity data: Application to function approximation and inverse pde problems.
\newblock \emph{Journal of Computational Physics}, 401:\penalty0 109020, 2020.

\bibitem[Pawar et~al.(2022)Pawar, Sharma, Vijayakumar, Bay, Yellapantula, and San]{pawar2022towards}
Suraj Pawar, Ashesh Sharma, Ganesh Vijayakumar, Chrstopher~J Bay, Shashank Yellapantula, and Omer San.
\newblock Towards multi-fidelity deep learning of wind turbine wakes.
\newblock \emph{Renewable Energy}, 200:\penalty0 867--879, 2022.

\bibitem[Peherstorfer et~al.(2018)Peherstorfer, Willcox, and Gunzburger]{peherstorfer2018survey}
Benjamin Peherstorfer, Karen Willcox, and Max Gunzburger.
\newblock Survey of multifidelity methods in uncertainty propagation, inference, and optimization.
\newblock \emph{Siam Review}, 60\penalty0 (3):\penalty0 550--591, 2018.

\bibitem[Pfaff et~al.(2020)Pfaff, Fortunato, Sanchez{-}Gonzalez, and Battaglia]{DBLP:journals/corr/abs-2010-03409}
Tobias Pfaff, Meire Fortunato, Alvaro Sanchez{-}Gonzalez, and Peter~W. Battaglia.
\newblock Learning mesh-based simulation with graph networks.
\newblock \emph{CoRR}, abs/2010.03409, 2020.
\newblock URL \url{https://arxiv.org/abs/2010.03409}.

\bibitem[Raissi et~al.(2017{\natexlab{a}})Raissi, Perdikaris, and Karniadakis]{raissi2017inferring}
Maziar Raissi, Paris Perdikaris, and George~Em Karniadakis.
\newblock Inferring solutions of differential equations using noisy multi-fidelity data.
\newblock \emph{Journal of Computational Physics}, 335:\penalty0 736--746, 2017{\natexlab{a}}.

\bibitem[Raissi et~al.(2017{\natexlab{b}})Raissi, Perdikaris, and Karniadakis]{raissi2017physics}
Maziar Raissi, Paris Perdikaris, and George~Em Karniadakis.
\newblock Physics informed deep learning (part i): Data-driven solutions of nonlinear partial differential equations.
\newblock \emph{arXiv preprint arXiv:1711.10561}, 2017{\natexlab{b}}.

\bibitem[Raissi et~al.(2019)Raissi, Perdikaris, and Karniadakis]{RAISSI2019686}
Maziar Raissi, Paris Perdikaris, and George~E Karniadakis.
\newblock Physics-informed neural networks: A deep learning framework for solving forward and inverse problems involving nonlinear partial differential equations.
\newblock \emph{Journal of Computational physics}, 378:\penalty0 686--707, 2019.

\bibitem[Sanchez-Gonzalez et~al.(2020)Sanchez-Gonzalez, Godwin, Pfaff, Ying, Leskovec, and Battaglia]{pmlr-v119-sanchez-gonzalez20a}
Alvaro Sanchez-Gonzalez, Jonathan Godwin, Tobias Pfaff, Rex Ying, Jure Leskovec, and Peter Battaglia.
\newblock Learning to simulate complex physics with graph networks.
\newblock In Hal~Daumé III and Aarti Singh, editors, \emph{Proceedings of the 37th International Conference on Machine Learning}, volume 119 of \emph{Proceedings of Machine Learning Research}, pages 8459--8468. PMLR, 7 2020.

\bibitem[Shlomi et~al.(2020)Shlomi, Battaglia, and Vlimant]{shlomi2020graph}
Jonathan Shlomi, Peter Battaglia, and Jean-Roch Vlimant.
\newblock Graph neural networks in particle physics.
\newblock \emph{Machine Learning: Science and Technology}, 2\penalty0 (2):\penalty0 021001, 2020.

\bibitem[Sirignano and Spiliopoulos(2018)]{sirignano2018dgm}
Justin Sirignano and Konstantinos Spiliopoulos.
\newblock Dgm: A deep learning algorithm for solving partial differential equations.
\newblock \emph{Journal of computational physics}, 375:\penalty0 1339--1364, 2018.

\bibitem[Taghizadeh et~al.(2024)Taghizadeh, Nabian, and Alemazkoor]{taghizadeh2024multifidelity}
Mehdi Taghizadeh, Mohammad~Amin Nabian, and Negin Alemazkoor.
\newblock Multifidelity graph neural networks for efficient and accurate mesh-based partial differential equations surrogate modeling.
\newblock \emph{Computer-Aided Civil and Infrastructure Engineering}, 2024.

\bibitem[Thuerey et~al.(2021)Thuerey, Holl, Mueller, Schnell, Trost, and Um]{thuerey2021pbdl}
Nils Thuerey, Philipp Holl, Maximilian Mueller, Patrick Schnell, Felix Trost, and Kiwon Um.
\newblock \emph{Physics-based Deep Learning}.
\newblock WWW, 2021.
\newblock URL \url{https://physicsbaseddeeplearning.org}.

\bibitem[Tompson et~al.(2016)Tompson, Schlachter, Sprechmann, and Perlin]{tompson2016accelerating}
Jonathan Tompson, Kristofer Schlachter, Pablo Sprechmann, and Ken Perlin.
\newblock Accelerating eulerian fluid simulation with convolutional networks.
\newblock \emph{arXiv preprint arXiv:1607.03597}, 2016.

\bibitem[Veli{\v{c}}kovi{\'c} et~al.(2017)Veli{\v{c}}kovi{\'c}, Cucurull, Casanova, Romero, Lio, and Bengio]{velivckovic2017graph}
Petar Veli{\v{c}}kovi{\'c}, Guillem Cucurull, Arantxa Casanova, Adriana Romero, Pietro Lio, and Yoshua Bengio.
\newblock Graph attention networks.
\newblock \emph{arXiv preprint arXiv:1710.10903}, 2017.

\bibitem[Wang et~al.(2021{\natexlab{a}})Wang, Planas, Chandramowlishwaran, and Bostanabad]{wang2021train}
Hengjie Wang, Robert Planas, Aparna Chandramowlishwaran, and Ramin Bostanabad.
\newblock Train once and use forever: Solving boundary value problems in unseen domains with pre-trained deep learning models.
\newblock \emph{arXiv e-prints}, pages arXiv--2104, 2021{\natexlab{a}}.

\bibitem[Wang et~al.(2021{\natexlab{b}})Wang, Teng, and Perdikaris]{wang2021understanding}
Sifan Wang, Yujun Teng, and Paris Perdikaris.
\newblock Understanding and mitigating gradient flow pathologies in physics-informed neural networks.
\newblock \emph{SIAM Journal on Scientific Computing}, 43\penalty0 (5):\penalty0 A3055--A3081, 2021{\natexlab{b}}.

\bibitem[Wei and Chen(2019)]{wei2019physics}
Zhun Wei and Xudong Chen.
\newblock Physics-inspired convolutional neural network for solving full-wave inverse scattering problems.
\newblock \emph{IEEE Transactions on Antennas and Propagation}, 67\penalty0 (9):\penalty0 6138--6148, 2019.

\bibitem[Yu et~al.(2018)]{yu2018deep}
Bing Yu et~al.
\newblock The deep ritz method: a deep learning-based numerical algorithm for solving variational problems.
\newblock \emph{Communications in Mathematics and Statistics}, 6\penalty0 (1):\penalty0 1--12, 2018.

\bibitem[Yun et~al.(2019)Yun, Jeong, Kim, Kang, and Kim]{yun2019graph}
Seongjun Yun, Minbyul Jeong, Raehyun Kim, Jaewoo Kang, and Hyunwoo~J Kim.
\newblock Graph transformer networks.
\newblock \emph{Advances in neural information processing systems}, 32, 2019.

\bibitem[Zhang and Garikipati(2020)]{zhang2020machine}
Xiaoxuan Zhang and Krishna Garikipati.
\newblock Machine learning materials physics: Multi-resolution neural networks learn the free energy and nonlinear elastic response of evolving microstructures.
\newblock \emph{Computer Methods in Applied Mechanics and Engineering}, 372:\penalty0 113362, 2020.

\bibitem[Zhang et~al.(2021)Zhang, Xie, Ji, Zhu, and Zheng]{zhang2021multi}
Xinshuai Zhang, Fangfang Xie, Tingwei Ji, Zaoxu Zhu, and Yao Zheng.
\newblock Multi-fidelity deep neural network surrogate model for aerodynamic shape optimization.
\newblock \emph{Computer Methods in Applied Mechanics and Engineering}, 373:\penalty0 113485, 2021.

\bibitem[Zhou et~al.(2020)Zhou, Cui, Hu, Zhang, Yang, Liu, Wang, Li, and Sun]{zhou2020graph}
Jie Zhou, Ganqu Cui, Shengding Hu, Zhengyan Zhang, Cheng Yang, Zhiyuan Liu, Lifeng Wang, Changcheng Li, and Maosong Sun.
\newblock Graph neural networks: A review of methods and applications.
\newblock \emph{AI Open}, 1:\penalty0 57--81, 2020.

\bibitem[Zhu et~al.(2019)Zhu, Zabaras, Koutsourelakis, and Perdikaris]{zhu2019physics}
Yinhao Zhu, Nicholas Zabaras, Phaedon-Stelios Koutsourelakis, and Paris Perdikaris.
\newblock Physics-constrained deep learning for high-dimensional surrogate modeling and uncertainty quantification without labeled data.
\newblock \emph{Journal of Computational Physics}, 394:\penalty0 56--81, 2019.

\end{thebibliography}
\medskip
\end{document}